\documentclass[times, review, 10pt]{elsarticle}

\usepackage{amssymb}
\usepackage{amsmath}
\usepackage{hyperref}  
\usepackage{multirow} 
\usepackage{graphicx} 
\usepackage{colortbl} 
\usepackage{xcolor}   
\usepackage{tabularx} 
\usepackage{graphicx}  
\usepackage{url} 
\usepackage{booktabs}
\usepackage{cleveref}
\usepackage{subcaption}
\usepackage{listings}

\journal{PATTERN RECOGNITION}

\begin{document}

\begin{frontmatter}



\title{PointSlice: Accurate and Efficient Slice-Based Representation for 3D Object Detection from Point Clouds}


\author{Liu Qifeng, Zhao Dawei, Dong Yabo, Xiao Liang, Wang Juan, Min Chen, Li Fuyang, Jiang Weizhong} 
\author{Lu Dongming\corref{cor1}}
\author{Nie Yiming}

\cortext[cor1]{Corresponding author. Email address: ldm@zju.edu.cn}
\affiliation{organization={Zhejiang University},
            addressline={College of Computer Science and Technology}, 
            city={Hangzhou, Zhejiang},
            postcode={310058}, 
            state={Zhejiang},
            country={China}}
\affiliation{organization={Defense Innovation Institute},
            city={Beijing},
            postcode={100000}, 
            state={Beijing},
            country={China}}
\affiliation{organization={Tsinghua University},
            addressline={Department of Computer Science and Technology}, 
            city={Beijing},
            postcode={100000}, 
            state={Beijing},
            country={China}}
\begin{abstract}
3D object detection from point clouds plays a critical role in autonomous driving. Currently, the primary methods for point cloud processing are voxel-based and pillar-based approaches. Voxel-based methods offer high accuracy through fine-grained spatial segmentation but suffer from slower inference speeds. Pillar-based methods enhance inference speed but typically lag behind voxel-based methods in detection accuracy. To address this trade-off, we propose a novel point cloud processing method, PointSlice, which slices point clouds along the horizontal plane and incorporates a dedicated detection network. The main contributions of PointSlice are: (1) A novel slice-based representation that converts 3D point clouds into multiple sets of 2D (x-y) data slices. The model explicitly learns 2D data distributions by treating the 3D point cloud as separate batches of 2D data, which significantly reduces the parameter count and enhances inference speed; (2) The introduction of a Slice Interaction Network (SIN). To preserve vertical geometric relationships across slices, we incorporate SIN into the 2D backbone network, thereby improving the model's 3D perception capability. Extensive experiments demonstrate that PointSlice achieves a superior balance between detection accuracy and efficiency. On the Waymo Open Dataset, PointSlice achieves a 1.13$\times$ speedup and uses 0.79$\times$ the parameters of the state-of-the-art voxel-based method (SAFDNet), with a marginal 1.2 mAPH accuracy reduction. On the nuScenes dataset, we achieve a state-of-the-art 66.7 mAP. On the Argoverse 2 dataset, PointSlice is 1.10$\times$ faster with 0.66$\times$ the parameters, while showing a negligible accuracy drop of 1.0 mAP. The source code is available at \url{https://github.com/qifeng22/PointSlice2}.
\end{abstract}




\end{frontmatter}



\section{ Introduction}
\label{sec:intro}

\begin{figure}[htbp] 
  \centering
  \includegraphics[width=0.75\linewidth,height=0.75\linewidth]{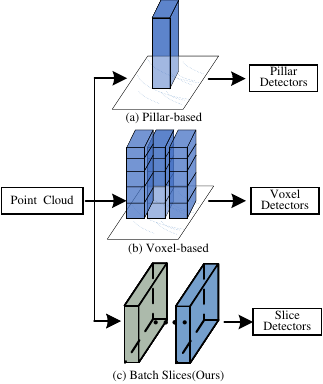} 
  \caption{Comparison of different point cloud processing methods: pillar-based, voxel-based, and batch slices (Ours).}
  \label{fig:voxelpillarcompare}
\end{figure}
LiDAR-based 3D object detection has been extensively studied due to its critical applications in autonomous driving and robotics~\cite{roboticszhu2017target}. 
Recently, significant progress has been made across various research directions. For instance, \cite{yin2022semi_r1} proposed a semi-supervised framework with proficient teachers to alleviate the reliance on massive annotations, while \cite{yan2025olidm_r2} explored diffusion models for object-aware data generation. Furthermore, the scope of detection has been extended to vehicle-infrastructure collaboration~\cite{li2024di_r3} and spatiotemporal video processing~\cite{yin2021graph_r4}. Despite these advancements in learning paradigms and application scenarios, the fundamental trade-off between detection accuracy and inference speed in the backbone architecture remains a critical bottleneck.

The inherent sparsity and uneven spatial distribution of point clouds make it challenging to directly apply 2D image network architectures to point cloud data. To address these characteristics, the mainstream approach for processing point clouds is voxelization. The two primary voxelization methods are voxel-based and pillar-based approaches (see \cref{fig:voxelpillarcompare}). Voxel-based methods divide the point cloud space into a 3D voxel grid; models such as HEDNet \cite{hednet} and SAFDNet \cite{safdnet} utilize hierarchical encoder-decoder architectures and sparse detection head structures, achieving outstanding detection performance. However, voxel-based approaches require learning in the x, y, and z dimensions, which results in slower inference speeds compared to pillar-based methods. Pillar-based methods compress the point cloud data into the x-y plane, reducing input dimensionality and thus improving inference efficiency. For instance, PillarNet \cite{pillarnet} employs spatial feature semantic fusion to further enhance detection accuracy. Given the efficiency advantages of the pillar-based approach, a natural question arises: can the voxel-based network structure be directly applied to point clouds initialized in a pillar format? We conducted experiments on the Waymo Open dataset to explore this question. As shown in \cref{SAFDNetpillar}, although using pillar-formatted point clouds (SAFD-Pillar) results in an inference speed 1.17× faster than SAFDNet, it achieves only 69.3 mAPH in detection accuracy.

\begin{table}[hp]
  \centering
  \begin{tabular}{@{}lcccc@{}}
    \toprule
    Method & mAPH(L2) & Params & Speedup & FPS  \\
    \midrule
    SAFDNet   & 73.9   & 9.89M   & 1.0$\times$  & 13.68    \\
    SAFD-Pillar   & 69.3   & 7.74M  & 1.17$\times$  & 15.95  \\
    \bottomrule
  \end{tabular}
  \caption{Comparison of pillarization results for SAFDNet\cite{safdnet} on the Waymo validation set. }
  \label{SAFDNetpillar}
\end{table}
To address the issue of improved efficiency but reduced detection accuracy, we propose PointSlice, which converts 3D point cloud data into multiple (x-y) slices, allowing the model to maintain detection accuracy close to that of voxel-based methods. PointSlice introduces a novel encoding approach: we partition the point cloud horizontally, transforming 3D data into a set of 2D slices, where the neural network does not differentiate between the heights (z positions) of the slices. This set of 2D slices is treated as a batch. Through this encoding, PointSlice can utilize a 2D convolutional network as the backbone for feature extraction, thereby overcoming the inefficiency of voxel-based methods that require 3D CNNs for inference.

However, due to the relationships among different slices of the same point cloud, solely learning from individual 2D slices would likely compromise the model’s final detection performance, as shown in the ablation studies. To address this, we further introduce the Slice Interaction Network (SIN), which incorporates sparse 3D convolutions within the 2D backbone network to enable information exchange across slices. Since SIN is composed of 3D convolutions, excessive use would increase model parameters and reduce inference efficiency. Therefore, we carefully add SIN only where necessary to preserve detection accuracy. Experimental results validate the importance of the SIN network.

To validate the effectiveness of our model, we conducted experiments on the Waymo and nuScenes datasets. On the Waymo dataset, our model achieves an inference speed that is $1.13 \times$ faster than the state-of-the-art voxel-based method (SAFDNet), with $0.79\times$ fewer parameters. On the nuScenes dataset, our model demonstrates $0.45\times$ fewer parameters than SAFDNet, while achieving a state-of-the-art detection result of 66.74 mAP.

Overall, the contributions of this paper are as follows: 

\textbf{(1)} we propose a novel representation that converts 3D point clouds into 2D slices; 

\textbf{(2)} we design a dedicated network structure, PointSlice, which innovatively incorporates the Slice Interaction Network (SIN); 

\textbf{(3)} we demonstrate the efficiency and high accuracy of our model on the Waymo, nuScenes, and Argoverse 2 datasets.

\section{Related work}
\label{sec:relatework}
\subsection{Voxel-based 3D detectors}
Voxel-based detection networks divide point cloud data into 3D voxel grids and utilize neural networks for 3D object detection. VoxelNet \cite{zhou2018voxelnet} divides the point cloud into evenly spaced 3D voxels and converts each point cluster within a voxel into a unified feature representation through the Voxel Feature Encoding (VFE) layer. Voxel R-CNN \cite{voxelrcnn} achieves precise object localization by fully leveraging voxel features within a two-stage approach. TransFusion \cite{bai2022transfusion} introduces a Transformer decoder-based detection head. DCT \cite{cui2025dynamic_pr} proposes a clustering-based Transformer backbone that exploits the inherent separability of LiDAR point clouds, employing Sparse Cluster Generation and Cluster Feature Interaction to facilitate dynamic information propagation between clusters and voxels. HEDNet \cite{hednet} employs encoder-decoder blocks to capture long-range dependencies between features. SVEFusion \cite{yang2025beyond} extends voxel-based representations to LiDAR and 4D radar fusion, employing a salient voxel enhancement mechanism to address the lack of fine-grained structural details in pillar-based methods. Beyond performance improvements, recent research has also investigated the robustness of these architectures. GSVA \cite{wu2025gradient} reveals the vulnerability of voxel-based detectors to adversarial attacks and proposes a gradient-based sparse voxel attack framework to evaluate their security boundaries. These methods have progressively improved detection accuracy and robustness, though inference efficiency remains a challenge.

To address the issue of inference efficiency, several works have explored solutions \cite{fullysparse,fsdv2}. VoxelNeXt\cite{voxelnext} introduced a fully sparse detection network with a sparse detection head, addressing the inefficiency in prior voxel-based networks that required sparse features to be converted into dense feature maps. This conversion reduced inference efficiency. SAFDNet\cite{safdnet} further improved the sparse detection head by designing an adaptive feature diffusion strategy to address the issue of missing central features for detected objects, achieving state-of-the-art detection results while ensuring high inference efficiency. However, even with fully sparse detectors, voxel-based methods still lag behind the upcoming pillar-based methods in terms of inference efficiency.

\subsection{Pillar-based 3D detectors}
Pillar-based methods \cite{pillarnext,pointpillars} build upon voxel-based approaches to further enhance the model's deployment feasibility in practical applications. PointPillars \cite{pointpillars} represents point clouds as vertical columns (pillars) and uses a 2D convolutional detection architecture for object detection. CenterPoint \cite{yin2021centerpoint} improves the accuracy of pillar-based methods by employing a keypoint detector to identify object centers. PillarNet \cite{pillarnet} further proposes a neck network with spatial semantic feature fusion. \cite{voxelorpillar} proposed a hybrid Voxel-Pillar Fusion network (VPF), which combines the strengths of voxel and pillar representations. They developed a sparse voxel-pillar encoder using 3D and 2D sparse convolutions and introduced a Sparse Fusion Layer (SFL) to enable bidirectional interaction between voxel and pillar features. Although these studies have continuously improved the detection accuracy of pillar-based methods, a gap remains when compared to the advancing voxel-based methods.

To address the limitations of both types of detectors, we propose a novel point cloud encoding method PointSlice, which slices the point cloud horizontally and employs a Slice Interaction Network (SIN) to enable efficient and accurate 3D object inference.

\section{PointSlice}
\label{sec:formatting}
\subsection{Background}
\textbf{Sparse convolutions.} Sparse convolution is a convolutional operation developed specifically for sparse data, operating only on locations where data exists. This approach enhances computational efficiency and reduces memory usage. Sparse convolution is mainly divided into two types: submanifold sparse convolution (subm spconv) \cite{graham2017submanifold}, which maintains feature sparsity between input and output feature maps, and regular sparse convolution \cite{regularsparseconv}, which increases feature map density by expanding features into neighboring areas, further filling the sparse space. In our Slice Interaction Network (SIN) module, we employ both types of convolution for information exchange between slices.\\
\textbf{Sparse encoder-decoder block (EDB).} HEDNet \cite{hednet} proposes merging sparse encoder-decoder blocks that capture long-range dependencies between features while maintaining computational efficiency.\\
\textbf{Sparse detection head.} Sparse detection heads primarily address the redundant computations associated with dense feature maps. SAFDNet \cite{safdnet} maintains feature sparsity throughout the entire detection process, while its proposed Adaptive Feature Diffusion (AFD) strategy mitigates the issue of missing central features, enabling fully sparse detection networks to achieve higher accuracy.

\begin{figure}[t] 
  \centering
  \includegraphics[width=1.0\linewidth]{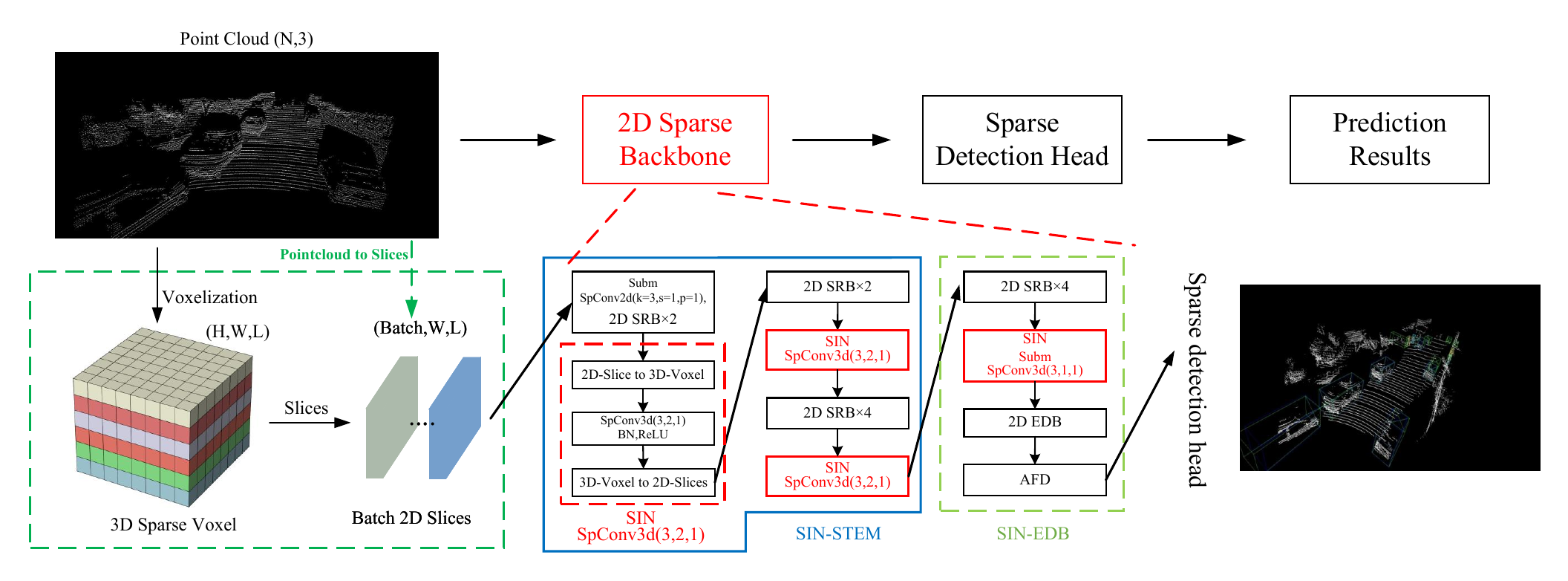} 
  \caption{Overall framework of PointSlice. The dashed boxes labeled ``Pointcloud to Slice'' and ``Slice Interaction Network (SIN)'' represent the main contributions of this paper. The raw point clouds, after voxelization and slicing, are input into a 2D backbone network for feature extraction. The 2D backbone network is composed of SIN-STEM and SIN-EDB. The SIN-STEM consists of 2D Sparse Residual Blocks (2DSRB) and SIN, which are responsible for efficient and effective feature generation. The SIN-EDB is constructed from 2DSRB, SIN, 2D Sparse encoder-decoder block (2DEDB), and AFD modules, designed to capture long-range dependencies among features.}
  \label{fig:architecture}
\end{figure}

\subsection{Overall architecture}
\cref{fig:architecture} illustrates the overall architecture of the PointSlice model, which operates in three stages: (1) transforming the point cloud into multiple sets of 2D slices; (2) extracting slice features through a sparse backbone, where the Slice Interaction Network (SIN) enables efficient information exchange across slices; and (3) performing final detection via a sparse detection head.
\subsubsection{Point cloud to Slices}
Given a set of point cloud data with a shape of ($B$, $N$, $3$), where $B$ denotes the batch size, $N$ represents the number of points in each point cloud (assuming each frame has an identical number of points), $3$ corresponds to the spatial coordinates $(x, y, z)$. Initially, we voxelize the point cloud following the approach of VoxelNet \cite{zhou2018voxelnet}, converting the point cloud into a regular voxel grid. We use \(H\), \(W\), and \(L\) to denote the height, width, and length of the entire voxel space, respectively. A batch of point clouds represented as voxels is denoted by \((B, H, W, L)\). This implies that each voxel coordinate lies within the range $(0, 0, 0, 0)$ to $(B-1, H-1, W-1, L-1)$. The feature of each voxel is generated using a PointNet \cite{qi2017pointnet} network, allowing the point cloud to be converted into a sparse tensor:
\begin{equation}
S=(\text{feature}, \text{indices}, \text{shape}, \text{batch}),
\label{yuanzhang3d}
\end{equation}
 where \(\text{feature}\) $(N^v,c)$ represents the voxel feature with $c$ feature channels, \(\text{indices}\) $(N^v,4)$ denotes the voxel coordinates \(\{(b_i, h_i, w_i, l_i)|\  i=0,...,N^v-1\}\), \(\text{shape}\) $(H,W,L)$ defines the dimensions of the entire voxel space, and \(N^v\) is the number of voxels. 

As seen, the voxel space is represented in 3D, which increases model parameters and inference time. Therefore, we simplify the voxel space by converting the 3D voxel space into \(H\) 2D slices, represented as sparse tensors:
\begin{equation}
S^{2D} = (\text{feature}, \text{indices}^{\text{2D}}, \text{shape}^{\text{2D}}, \text{batch} \times H),
\end{equation}
 where \(H \) represents the height of the original 3D voxel space, a parameter needed later in the Slice Interaction Network (SIN), $\text{shape}^{\text{2D}}$ $(W,L)$ represents the dimensions of a 2D slice. The $i$-th voxel coordinates for \(\text{indices}^{\text{2D}}\) $(N^v, 3)$ are defined as \((b^{2D}_i, w^{2D}_i, l^{2D}_i)\), with the following transformation:
\begin{equation}
b^{2D}_i = b_i \times H + h_i, \quad w^{2D}_i = w_i, \quad l^{2D}_i = l_i.
\label{eq:example}
\end{equation}
 Notably, since we only combine the original voxel space in the height direction (z-axis) into the batch dimension, the number of voxels \(N^v\) remains unchanged. In \eqref{eq:example}, the height of the 3D tensor is converted into the batch dimension. This slicing operation reduces the 3D point cloud representation into a 2D sparse tensor, thus reducing the data dimensionality by one. Consequently, model computation is parallelized in 2D, which reduces both the number of model parameters and inference time.
 
 \begin{figure}[t] 
  \centering
  \includegraphics[width=0.75\linewidth]{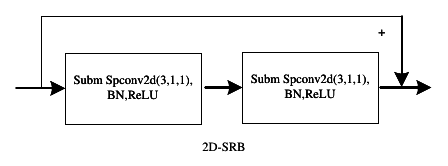} 
  \caption{Detailed structure of the 2D-SRB.}
  \label{fig:2dsrb}
\end{figure}
\subsubsection{Sparse 2D backbone}
\begin{figure}[htbp] 
  \centering
  \includegraphics[width=0.8\linewidth]{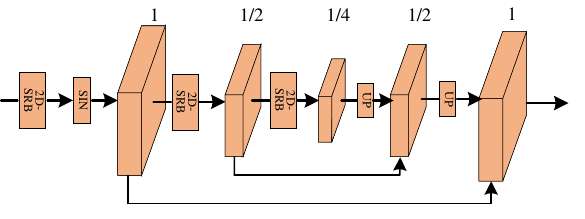} 
  \caption{Composition of the SIN and 2D-EDB models, with the design of the 2D-EDB module adapted from SAFDNet \cite{safdnet}.}
  \label{fig:2dedb}
\end{figure}

For the transformed slice tensor, we designed a 2D feature extraction network as the backbone of the detector. This backbone network is inspired by the structure of the SAFDNet\cite{safdnet} model and is composed of 2D Sparse Residual Blocks (2D-SRB) (\cref{fig:2dsrb}), 2D Sparse Encoder-Decoder Blocks (2D-EDB) (\cref{fig:2dedb}), and Adaptive Feature Diffusion (AFD). The 2D-SRB module includes a series of sparse residual blocks, each containing two 2D submanifold sparse convolutions and a skip connection linking its input and output. For the 2D-EDB module, regular sparse convolutions with a stride of 2 are used for downsampling the feature map, while sparse transposed convolutions \cite{spconv2022} are applied for upsampling. The entire encoding-decoding process utilizes 2D kernels. After feature extraction, the AFD structure diffuses sparse feature maps to improve the detection accuracy of the sparse detection head.

\subsubsection{Slice Interaction Network}
Converting point clouds into slices, if processed solely as 2D slices, inevitably loses inter-slice connections, specifically the height information critical for 3D objects. As shown in \cref{SINablation}, the results indicate that the lack of inter-slice interaction leads to a decrease in accuracy. To address this, we propose the Slice Interaction Network (SIN). The core idea is to assemble multiple slices of the same point cloud into 3D voxels, enabling inter-slice interaction through 3D regular sparse convolutions and submanifold sparse convolutions.

Before inputting multiple slices into the network, these slices need to be transformed into 3D voxels, a process represented by the following formula, Given a batch of 2D slice \(S^{2D}\), the corresponding 3D voxel representation is \cref{yuanzhang3d}.
The coordinate of the \(i\)-th voxel is:
\begin{equation}
indices_i = \left(b^{2D}_i \left\lfloor \frac{1}{H} \right\rfloor, b^{2D}_i \bmod H, w_i, l_i\right).
\end{equation} 
Since regular sparse 3D convolution (spconv3d) has demonstrated strong performance and simplicity in prior research \cite{safdnet,hednet}, we incorporate three spconv3d modules with kernel size = 3 and stride = 2 as SIN modules in the 2D-SRB stage. In the 2D-EDB stage, to enhance computational efficiency and ensure that input and output feature maps have identical dimensions, we use submanifold sparse 3D convolutions (subm spconv3d) with kernel size = 3 and stride = 1 for SIN in this stage. Overall, for the SIN module design, we select the straightforward structures of regular spconv3d and subm spconv3d, which effectively facilitate information exchange across slices, thus preserving the height information essential for 3D objects.

\subsubsection{Theoretical Complexity Analysis}
\label{sec:complexity_analysis}

The core innovation of PointSlice lies in its slice-based processing paradigm, which transforms the computationally intensive 3D voxel feature extraction into efficient 2D slice processing. As illustrated in \cref{fig:architecture}, the computational overhead is concentrated in the 2D Sparse Backbone, comprising the 2D-SRB, 2D-EDB, and SIN modules.

To quantify the computational complexity, we decompose the backbone into $m$ sparse 2D convolution layers and $n$ sparse 3D convolution layers (used for SIN interaction). Based on our model configuration for the Waymo dataset, we set $m=33$ and $n=4$. We define the theoretical computational cost (FLOPs) for a single sparse 2D and 3D convolution layer as follows:
\begin{equation}
    \mathcal{O}_{Sparse2D} \approx N_{act} \cdot C_{in} \cdot C_{out} \cdot K^2
\end{equation}
\begin{equation}
    \mathcal{O}_{Sparse3D} \approx N_{act} \cdot C_{in} \cdot C_{out} \cdot K^3
\end{equation}
where bias terms and activation functions are omitted for simplicity. Consequently, the total computational complexity of the PointSlice backbone, $\mathcal{O}_{Ours}$, can be expressed as:
\begin{equation}
    \mathcal{O}_{Ours} \approx \sum_{i=1}^{m} \mathcal{O}_{Sparse2D}^{(i)} + \sum_{j=1}^{n} \mathcal{O}_{Sparse3D}^{(j)}
\end{equation}
Assuming the channel dimensions $C$ ($C \approx C_{in} \approx C_{out}$) and the number of active voxels $N_{act}$ are of the same magnitude across layers, the formula can be approximated as:
\begin{equation}
    \mathcal{O}_{Ours} \approx N_{act} \cdot C^2 \cdot (m \cdot K^2 + n \cdot K^3)
\end{equation}
Here, the variables are defined as follows:
\begin{itemize}
    \item $N_{act}$: The number of active voxels/pixels. Since sparse convolution only computes at non-empty positions, this value is significantly smaller than the total spatial volume.
    \item $C$: The number of feature channels.
    \item $K$: The kernel size, set to $3$ in our implementation.
    \item $m$: The number of sparse 2D convolution layers ($m=33$), primarily used for efficient feature extraction.
    \item $n$: The number of sparse 3D convolution layers ($n=4$), exclusively used for slice interaction in the SIN module.
\end{itemize}

In comparison, the backbone of SAFDNet consists entirely of 36 sparse 3D convolution layers. Its theoretical complexity is:
\begin{equation}
    \mathcal{O}_{SAFD} \approx 36 \cdot (N_{act} \cdot C^2 \cdot K^3)
\end{equation}

\noindent\textbf{Theoretical Comparison Results.}
Comparing $\mathcal{O}_{Ours}$ and $\mathcal{O}_{SAFD}$, the efficiency advantage of PointSlice is evident. SAFDNet incurs expensive cubic computational costs ($K^3=27$) across all 36 layers. In contrast, PointSlice retains only a minimal number of 3D convolutions ($n=4$) for necessary inter-slice interaction, while the vast majority of layers ($m=33$) utilize efficient quadratic 2D convolutions ($K^2=9$).

It is worth noting that theoretical FLOPs do not fully reflect actual inference latency. As discussed in SECOND~\cite{second}, sparse convolution involves not only numerical multiplication but also \textit{Rulebook Generation} and \textit{Hash Table Lookup} to establish input-output index mappings. This process introduces significant irregular memory access overhead. By drastically reducing the number of expensive sparse 3D convolution layers from 36 to 4, PointSlice effectively minimizes this memory access latency. To validate this analysis, we benchmarked the backbone inference time on the Waymo dataset. As shown in \cref{tab:inference_speed}, PointSlice saves approximately \textbf{7ms} in backbone computation per frame compared to SAFDNet, achieving highly efficient object detection.

\subsubsection{Sparse detection head}
In this work, we adopt the efficient sparse detection head as the final output layer. The 2D slice-based backbone, SIN module, and sparse detection head collectively constitute PointSlice. Through the transformation of point clouds into 2D slices, inter-slice information exchange, and a fully sparse detector, PointSlice demonstrates both high efficiency and precision in performance.

\begin{table*}
  \centering
   
\resizebox{\textwidth}{!}{%
  \begin{tabular}{@{}l|c|c|c|c|c|c|c|c@{}}
    \toprule
    \multirow{2}{*}{Method}  & \multicolumn{2}{c|}{ Vehicle AP/APH} & \multicolumn{2}{c|}{Pedestrian AP/APH} & \multicolumn{2}{c|}{Cyclist AP/APH} & \multirow{2}{*}{FPS} &\multirow{2}{*}{Params} \\
                              & L1   & L2   & L1   & L2   & L1   & L2   &     \\
    \midrule
     PointPillar\cite{pointpillars}   & 72.1/71.5 &63.6/63.1  & 70.6/56.7 & 62.8/50.3 & 64.4/62.3 & 61.9/59.9 & 27.1&4.84M \\
    CenterPoint\cite{yin2021centerpoint}   & 73.4/72.9 & 65.1/64.6 & 75.4/65.1 & 67.6/58.3 & 67.8/66.2 & 65.3/63.8& 25.8 &5.22M  \\
    PillarNeXt\cite{pillarnext}   &  78.4/77.9 & 70.3/69.8 & 82.5/77.1 & 74.9/69.8 & 73.2/72.2 & 70.6/69.6 & 7.7  &9.47M\\
    Voxel-or-Pillar\cite{voxelorpillar}    &  80.2/79.7 & 71.9/71.5 & 82.5/76.9 & 74.8/69.4 & 77.1/76.0 & 74.2/73.2 & 17.2 &-\\
    PillarNet-34\cite{pillarnet}  & 79.1/78.6  & 70.9/70.5& 80.6/74.0 & 72.3/66.2 & 72.3/71.2 & 69.7/68.7 & - & - \\
    PillarNet-18\cite{pillarnet}  & 78.2/77.7  & 70.4/69.9& 79.8/72.6 & 71.6/64.9 & 70.4/69.3 & 67.8/66.7 & 22.7 & 13.42M \\
    
    \midrule
    SECOND\cite{second}   & 72.3/71.7 & 63.9/63.3& 68.7/58.2 & 60.7/51.3 & 60.6/59.3 & 58.3/57.0 & 31.4&5.33M \\
    Part-A2\cite{shi2020points}   & 77.1/76.5 & 68.5/68.0 & 75.2/66.9 & 66.2/58.6 & 68.6/67.4 & 66.1/64.9 & 7.5 &39.96M\\
    PV-RCNN\cite{pvrcnn}   & 78.0/77.5 & 69.4/69.0 & 79.2/73.0 & 70.4/64.7 & 71.5/70.3 & 69.0/67.8 & 3.1 &13.50M \\
    AFDetV2\cite{afdetv2}  & 77.6/77.1 & 69.7/69.2 & 80.2/74.6 & 72.2/67.0 & 73.7/72.7 & 71.0/70.1 & 22.2 &- \\
    CenterFormer\cite{zhou2022centerformer} &75.0/74.4&69.9/69.4&78.0/72.4& 73.1/67.7&73.8/72.7&71.3/70.2&-&-\\
    PV-RCNN++\cite{pvrcnn++}  & 79.3/78.8 & 70.6/70.2 & 81.3/76.3 & 73.2/68.0 & 73.7/72.7 & 71.2/70.2 &7.4&16.10M\\
    DSVT-Voxel\cite{dsvt}&79.7/79.3&71.4/71.0&83.7/78.9&76.1/71.5& 77.5/76.5&74.6/73.7&8.4&9.10M\\
    SWFormer\cite{swformer} & 77.8/77.3& 69.2/68.8& 80.9/72.7& 72.5/64.9& -/-&-/-&-&-\\
    VoxelNeXt\cite{voxelnext}&78.2/77.7&69.9/69.4&81.5/76.3&73.5/68.6&76.1/74.9&73.3/72.2&7.9&34.01M\\
    PTv3\cite{pvt3} &-/- &71.2/70.8&-/-&76.3/70.4&-/-&71.5/70.4&12.5&5.23M\\
    FSDV1\cite{fullysparse}&79.2/78.8&70.5/70.1&82.6/77.3&73.9/69.1&77.1/76.0&74.4/73.3&7.1&24.31M\\
    FSDv2\cite{fsdv2}&79.8/79.3&71.4/71.0&84.8/79.7&77.4/72.5&80.7/79.6&77.9/76.8&9.5&44.83M\\
    HEDNet\cite{hednet}&81.1/80.6&73.2/72.7&84.4/80.0&76.8/72.6&78.7/77.7&75.8/74.9&13.2&12.64M\\
    
    Voxel Mamba \cite{zhang2024voxelmamba} &  80.8/80.3 & 72.6/72.2 & 85.0/80.8 & 77.7/73.6 & 78.6/77.6 & 75.7/74.8 &8.4 &20.99M\\
    LION \cite{liu2024lion} & 80.3/79.9 & 72.0/71.6 & 85.8/81.4 & 78.5/74.3 & 80.1/79.0 & 77.2/76.2 & 4.9& 16.07M\\
    SAFDNet\cite{safdnet}&80.6/80.1&72.7/72.3&84.7/80.4&77.3/73.1&80.0/79.0&77.2/76.2&13.6&9.89M\\
    \midrule
    SAFDNet$^{\dag}$&80.6/80.2&72.7/72.2&85.0/80.7&77.7/73.4&79.3/78.2&76.4/75.4&13.6&9.89M\\
    \rowcolor{cyan!20} 
    PointSlice-W (Ours) &80.3/79.8&72.3/71.9&84.1/79.4&76.4/71.9&77.9/76.9&75.1/74.2&15.4&7.82M\\
    \rowcolor{cyan!20} 
    PointSlice-W-conv5 &80.4/80.0&72.5/72.1&84.3/79.7&76.8/72.4&78.3/77.3&75.5/74.5&14.6&8.43M\\

    \bottomrule
  \end{tabular}%
  }
  
  \caption{Comparison of performance between pillar-based and voxel-based models on the Waymo validation set, where $^{\dag}$ indicates our reproduced results. FPS (frame per second) was measured on a single NVIDIA 3090 GPU with a batch size of 1. All models were trained under single-frame setting.}
  \label{waymovalidation}
\end{table*}

\begin{table}[ht]
  \centering
  \resizebox{0.5\textwidth}{!}{%
    \begin{tabular}{@{}l|c@{}}
      \toprule
      Method & Backbone Latency (ms)  \\
      \midrule
      SAFDNet~\cite{safdnet} & 29 \\
      PointSlice-W (Ours) & \textbf{22}  \\
      \bottomrule
    \end{tabular}%
  }
  \caption{Comparison of backbone inference latency on the Waymo Open Dataset. The latency is measured in milliseconds (ms) on a single NVIDIA 3090 GPU with a batch size of 1.}
  \label{tab:inference_speed}
\end{table}

\section{Experiments}
\subsection{Datasets and Metrics}
In this work, we evaluate our proposed method on three large-scale autonomous driving datasets: Waymo Open Dataset \cite{waymosun2020scalability}, nuScenes \cite{nuscenes}, and Argoverse 2 \cite{argoverse}. The details of these datasets are described below:

\vspace{0.3cm} 

\noindent \textbf{Waymo Open Dataset (WOD):} WOD is recognized for its high-density point clouds and complex urban traffic scenarios. The dataset comprises 1,150 sequences (798 training, 202 validation, 150 testing), collected across diverse geographies including San Francisco, Phoenix, and Mountain View, covering various weather conditions. Captured by a high-quality sensor suite including five LiDARs and five cameras, the data is annotated at $10\,\text{Hz}$. The evaluation metrics are rigorous, employing 3D mean Average Precision (mAP) and mAP weighted by heading accuracy (mAPH). Specifically, the evaluation is stratified into two difficulty levels based on point density: Level 1 (L1) for objects containing more than five points, and Level 2 (L2) for challenging cases with at least one point, testing the model's ability to detect sparse objects.

\vspace{0.3cm} 

\noindent \textbf{nuScenes Dataset:} This dataset emphasizes multimodal sensor fusion and diverse driving conditions. It contains 1,000 scenes of 20 seconds each, collected in Boston and Singapore, which introduces unique challenges such as left-hand versus right-hand traffic and varying vegetation. Unlike WOD, nuScenes provides annotations at $2\,\text{Hz}$. The sensor suite includes one 32-beam LiDAR, five radars, and six cameras providing $360^{\circ}$ coverage. For evaluation, apart from the standard mAP (calculated based on center distance rather than IoU), nuScenes introduces a composite metric named nuScenes Detection Score (NDS). NDS encapsulates detection accuracy along with box location, size, orientation, attribute, and velocity errors, offering a holistic assessment of 3D perception performance.

\vspace{0.3cm} 

\noindent \textbf{Argoverse 2 Sensor Dataset (AV2):} As a next-generation benchmark, AV2 significantly expands the complexity and diversity of autonomous driving data. It consists of 1,000 scenes captured in six distinct U.S. cities (Austin, Detroit, Miami, Palo Alto, Pittsburgh, and Washington D.C.), ensuring a wide variance in urban layouts and object types. A key feature of AV2 is its focus on the long-tail distribution of objects, providing a comprehensive taxonomy of 30 classes (e.g., strollers, wheelchairs, construction equipment) to test the limits of semantic understanding. The dataset is equipped with high-resolution sensors (two 32-beam LiDARs and seven ring cameras) and includes rich HD Map priors, such as 3D lane geometry and ground height maps. We follow the standard protocol evaluating on 26 categories using mAP and composite errors similar to nuScenes.

\subsection{Data Augmentation}

To enhance the model's generalization capability and mitigate overfitting, we implement a comprehensive data augmentation strategy during training. Specifically, we employ the ``copy-paste'' database sampling technique, where ground truth objects from other frames are randomly injected into the current scene; to ensure sample quality, we filter out objects containing fewer than 5 points (raised to 10 points for pedestrians and cyclists in Waymo). Furthermore, we apply global geometric transformations, including random flipping along both the $X$ and $Y$ axes and global rotation with angles uniformly sampled from $[-\frac{\pi}{4}, \frac{\pi}{4}]$. The entire point cloud undergoes random scaling with factors sampled from $[0.95, 1.05]$ for Waymo and Argoverse 2, and $[0.9, 1.1]$ for nuScenes, followed by random global translation with Gaussian noise ($\sigma=0.5$) along all three axes. Finally, the input points are randomly shuffled to ensure invariance to permutation.

\subsection{Implementation details}

\noindent\textbf{Network Architecture.}
Our method is implemented based on OpenPCDet~\cite{openpcdet2020}.\\ PointSlice utilizes 8 2D sparse residual blocks, with SIN modules inserted after the 2nd, 4th, and 8th residual blocks. Each SIN module is composed of a Sparse Convolution with kernel size 3 and stride 2, configured with channel dimensions of 16$\to$32, 32$\to$64, and 64$\to$64, respectively. The model also includes a single 2D sparse EDB layer, within which one SIN module is inserted; this module consists of a Submanifold Sparse Convolution with kernel size 3 and stride 1, utilizing constant 64$\to$64 channel dimensions to maintain spatial resolution. 
In our implementation, we employ the standard \texttt{nn.BatchNorm1d} for feature normalization. Despite mapping the height dimension to the batch dimension to increase the effective batch size, our empirical results confirm that this standard formulation ensures stable training convergence and robust generalization without requiring special statistical alignment.
For the AFD and sparse detection head, the model adopts the same configuration as SAFDNet. Notably, for detection on the nuScenes dataset, the model employs the SparseTransFusionHead from TransFusion~\cite{bai2022transfusion} as its detection head.
Note that the PointSlice architecture is adapted for each benchmark to handle varying category counts and scene complexities. For clarity, we denote the specific variants as \textbf{PointSlice-N} (nuScenes), \textbf{PointSlice-W} (Waymo), and \textbf{PointSlice-A} (Argoverse 2).

\noindent\textbf{Training Settings.}
All experiments were conducted on 8 RTX 3090 GPUs with a total batch size of 16. For fair comparison, we trained PointSlice for 24 epochs, 20 epochs, and 24 epochs on the Waymo Open, nuScenes, and Argoverse 2 datasets, respectively. 
For the \textbf{Waymo Open dataset}, we set the voxel size to $(0.08\text{m}, 0.08\text{m}, 0.15\text{m})$ and the detection range to $[-75.52\text{m}, 75.52\text{m}]$ in the X and Y axes, and $[-2\text{m}, 4\text{m}]$ in the Z axis. We employed the Adam~\cite{adam} optimizer with a one-cycle learning rate policy, setting the weight-decay to 0.05 and the max learning rate to 0.003. 
For the \textbf{nuScenes dataset}, the voxel size was set to $(0.3\text{m}, 0.3\text{m}, 1.0\text{m})$ with the detection range of $[-54\text{m}, 54\text{m}]$ in the X and Y axes, and $[-5\text{m}, 3\text{m}]$ in the Z axis. 
For the \textbf{Argoverse 2 dataset}, the voxel size was $(0.1\text{m}, 0.1\text{m}, 0.2\text{m})$ and the detection range was $[-200\text{m}, 200\text{m}]$ in the X and Y axes, and $[-4\text{m}, 4\text{m}]$ in the Z axis.

\subsection{Comparison with state-of-the-art methods}
\noindent

\noindent \textbf{Results on Waymo Open Dataset.} \cref{waymovalidation} presents the performance results of the model on the validation set. PointSlice achieves an L2 mAPH of 72.7, an improvement of +5.5 over the state-of-the-art pillar-based method, PillarNet. 
Compared to the leading voxel-based method, SAFDNet, our model achieves a 1.13$\times$ speedup and requires only 0.79$\times$ the parameters, with only a 1.2 decrease in L2 mAPH.
 Furthermore, we extend the comparison to recent state-of-the-art methods. PointSlice achieves a remarkable 3.14$\times$ speedup compared to LION \cite{liu2024lion}. Against the efficient architecture PTv3 \cite{pvt3}, our model demonstrates superior performance in both detection accuracy and inference speed. Although our model exhibits a slight gap in inference speed compared to PillarNet, our approach further narrows the performance gap between voxel-based and pillar-based methods, providing new insights for future research.

\begin{table*}
  \centering
  \resizebox{\textwidth}{!}{%
    \begin{tabular}{l|cc|cccccccccccc}
      \hline
      Method & NDS & mAP  & Car & Truck  & Bus & T.L. & C.V. & Ped. & M.T. & Bike & T.C. & B.R.&FPS&Params \\
      \midrule
      \multicolumn{15}{c}{\textit{Results on the validation dataset}} \\
      \midrule
      PillarNet-18\cite{pillarnet}&67.4&59.9&-&-&-&-&-&-&-&-&-&-&10.3&14.57M \\
      VoxelNeXt\cite{voxelnext} & 68.7 & 63.5 & 83.9 & 55.5 & 70.5 & 38.1 & 21.1 & 84.6 & 62.8 & 50.0 & 69.4 & 69.4&7.9&7.98M \\
      FSDv2\cite{fsdv2} & 70.4 & 64.7 & 84.4 & 57.3 & 75.9 & 44.1 & 28.5 & 86.9 & 69.5 & 57.4 & 72.9 & 73.6&8.27&42.25M \\
      SAFDNet\cite{safdnet} & 71.0 & 66.3 & 87.6 & 60.8 & 78.0 & 43.5 & 26.6 & 87.8 & 75.5 & 58.0 & 75.0 & 69.7&6.4&15.74M \\
      \rowcolor{cyan!20} 
      PointSlice-N (Ours) & 70.9 & \textbf{66.7} & 87.9 & 63.8 & 78.3 & 47.8 & 26.0 & 88.1 & 73.7 & 58.0 & 75.8 & 68.1 &6.9&7.23M\\
      \midrule
      \multicolumn{15}{c}{\textit{Results on the test dataset}} \\
      \midrule
      PointPillars\cite{pointpillars} & 45.3 & 30.5 & 68.4 & 23.0 & 28.2 & 23.4 & 4.1 & 59.7 & 27.4 & 1.1 & 30.8 & 38.9&11.6&5.99M \\
      3DSSD\cite{3dssd} & 56.4 & 42.6 & 81.2 & 47.2 & 61.4 & 30.5 & 12.6 & 70.2 & 36.0 & 8.6 & 31.1 & 47.9 &-&-\\
      CenterPoint\cite{yin2021centerpoint} & 65.5 & 58.0 & 84.6 & 51.0 & 60.2 & 53.2 & 17.5 & 83.4 & 53.7 & 28.7 & 76.7 & 70.9&8.4&8.94M \\
      AFDetV2\cite{afdetv2} & 68.5 & 62.4 & 86.3 & 54.2 & 62.5 & 58.9 & 26.7 & 85.8 & 63.8 & 34.3 & 80.1 & 71.0 &-&-\\
      PillarNet-18\cite{pillarnet} & 70.8 & 65.0 & 87.4 & 56.7 & 60.9 & 61.8 & 30.4 & 87.2 & 67.4 & 40.3 & 82.1 & 76.0 &10.3&14.57M\\
      PillarNet-34\cite{pillarnet} & 71.4 & 66.0 & 87.6 & 57.5 & 63.6 & 63.1 & 27.9 & 87.3 & 70.1 & 42.3 & 83.3 & 77.2 &10.2&17.10M\\
      VoxelNeXt\cite{voxelnext} & 70.0 & 64.5 & 84.6 & 53.0 & 64.7 & 55.8 & 28.7 & 85.8 & 73.2 & 45.7 & 79.0 & 74.6 &7.9&7.98M\\
      TransFusion-L\cite{bai2022transfusion} & 70.2 & 65.5 & 86.2 & 56.7 & 66.3 & 58.8 & 28.2 & 86.1 & 68.3 & 44.2 & 82.0 & 78.2 &8.6&8.31M\\
      FSDv2\cite{fsdv2} & 71.7 & 66.2 & 83.7 & 51.6 & 66.4 & 59.1 & 32.5 & 87.1 & 71.4 & 51.7 & 80.3 & 78.7&8.27&42.25M \\
      Voxel-or-Pillar\cite{voxelorpillar} & 72.7& 67.0 & 85.8 & 55.1 & 63.5 & 62.1 & 33.3 & 87.6 & 72.5 & 48.6 & 82.9 & 78.2&-&- \\
      \rowcolor{cyan!20} 
      PointSlice-N (Ours) & 71.6 & \textbf{67.6} & 87.3 & 51.7 & 67.6 & 65.4 & 37.6 & 88.7 & 73.4 & 41.5 & 85.4 & 77.5 &6.9&7.23M\\
      \hline
    \end{tabular}%
  }
  \caption{Comparison of results for different models on the nuScenes dataset. ‘T.L.’, ‘C.V.’, ‘Ped.’, ‘M.T.’, ‘T.C.’, and ’B.R.’ denote trailer, construction vehicle, pedestrian, motor, traffic cone, and barrier.}
  \label{nuscenes}
\end{table*}

Additionally, according to the method proposed in \cite{pillarnext}, increasing the convolution kernel size can improve model accuracy. Inspired by this, we further expanded the subm3d convolution kernel size in the 2D-SRB and 2D-EDB modules from 3 to 5. As shown in \cref{waymovalidation}, the experimental results demonstrate that this adjustment improved the model's accuracy by 0.3 L2 mAPH while maintaining an inference speed of 14.6 FPS.

Finally, we analyze the class-specific performance. We observe that the performance drop compared to heavy voxel-based baselines is primarily concentrated in the Pedestrian and Cyclist categories. This can be attributed to the geometric sensitivity of these vertically elongated objects to the slicing operation, combined with their inherent point sparsity. However, this limitation is effectively mitigated by the scalability of our architecture. As demonstrated in our Scalability Analysis (see ~\cref{scalabity testing}), increasing the depth of slice interaction (e.g., stacking additional 3D SRBs) significantly enhances the feature extraction capability for these small objects, successfully narrowing the accuracy gap while retaining a competitive efficiency advantage.

\noindent
\textbf{Results on nuScenes.} As shown in \cref{nuscenes}, PointSlice achieves state-of-the-art results on the nuScenes validation set, with a mAP of 66.7. As indicated in \cref{nuscejian}, in terms of parameter count, PointSlice has 0.45× fewer parameters than SAFDNet and achieves a 1.08× faster inference speed.
On the nuScenes test dataset, our model achieved an outstanding performance of 67.6 mAP, demonstrating higher detection accuracy compared to the efficient models PillarNet-18\cite{pillarnet} and \cite{voxelorpillar}.

\begin{table}
  \centering
  \begin{tabular}{@{}lcccc@{}}
    \toprule
    Method & mAP & FPS & Params &Speedup \\
    \midrule
    SAFDNet\cite{safdnet} & 66.3 & 6.4&15.74M&1.0$\times$\\
     PointSlice-N (Ours)& 66.7&6.9 &7.23M&1.08$\times$ \\
    \bottomrule
  \end{tabular}
  \caption{Performance comparison of different models on the nuScenes validation set.}
  \label{nuscejian}
\end{table}

\begin{table*}
  \centering
  \resizebox{\textwidth}{!}{%
  \begin{tabular}{@{}l@{}|@{\hspace{1pt}}c@{\hspace{1pt}}|@{}r@{}*{25}{@{\hspace{3pt}}r@{\hspace{1pt}}}} 
    \toprule
    Method & \rotatebox{90}{mAP} & \rotatebox{90}{Vehicle} & \rotatebox{90}{Bus} & \rotatebox{90}{Pedestrian} & \rotatebox{90}{Stop Sign} & \rotatebox{90}{Box Truck} & \rotatebox{90}{Bollard} & \rotatebox{90}{C-Barrel} & \rotatebox{90}{Motorcyclist} & \rotatebox{90}{MPC-Sign} & \rotatebox{90}{Motorcycle} & \rotatebox{90}{Bicycle} & \rotatebox{90}{A-Bus} & \rotatebox{90}{School Bus} & \rotatebox{90}{Truck Cab} & \rotatebox{90}{C-Cone} & \rotatebox{90}{V-Trailer} & \rotatebox{90}{Sign} & \rotatebox{90}{Large Vehicle} & \rotatebox{90}{Stroller} & \rotatebox{90}{Bicyclist} & \rotatebox{90}{Truck} & \rotatebox{90}{MBT} & \rotatebox{90}{Dog} & \rotatebox{90}{Wheelchair} & \rotatebox{90}{W-Device} & \rotatebox{90}{W-Rider} \\
    \midrule
    \small CenterPoint\cite{yin2021centerpoint} & \small 22.0& \small 67.6 & \small38.9& \small 46.5 & \small 16.9 & \small 37.4 & \small 40.1 & \small 32.2 & \small 28.6 & \small  27.4 &  \small 33.4 & \small 24.5 & \small 8.7 & \small  25.8 & \small 22.6 & \small 29.5 & \small 22.4 & \small 6.3 & \small 3.9 & \small 0.5 & \small 20.1 & \small 22.1 & \small 0.0 & \small 3.9 & \small 0.5 & \small 10.9 & \small 4.2 \\
    \small HEDNet\cite{hednet} & \small 37.1& \small 78.2 & \small47.7& \small 67.6 & \small 46.4 & \small 45.9 & \small 56.9 & \small 67.0 & \small 48.7 & \small  46.5 &  \small 58.2 & \small 47.5 & \small 23.3 & \small  40.9 & \small 27.5 & \small 46.8 & \small 27.9 & \small 20.6 & \small 6.9 & \small 27.2 & \small 38.7 & \small 21.6 & \small 0.0 & \small 30.7 & \small 9.5 & \small 28.5 & \small 8.7 \\
    \small SAFDNet\cite{safdnet} & \small 39.7& \small 78.5 & \small49.4& \small 70.7 & \small 51.5 & \small 44.7 & \small 65.7 & \small 72.3 & \small 54.3 & \small  49.7 &  \small 60.8 & \small 50.0 & \small 31.3 & \small  44.9 & \small 24.7 & \small 55.4 & \small 31.4 & \small 22.1 & \small 7.1 & \small 31.1 & \small 42.7 & \small 23.6 & \small 0.0 & \small 26.1 & \small 1.4 & \small 30.2 & \small 11.5 \\
    \midrule
    
    \small PointSlice-A (Ours) & \small 38.7& \small 77.7 & \small47.8& \small 69.8 & \small 47.8 & \small 46.0 & \small 63.6 & \small 71.1 & \small 50.5 & \small  50.2 &  \small 59.6 & \small 50.5 & \small 28.9 & \small  44.4 & \small 26.1 & \small 54.8 & \small 31.9 & \small 20.8 & \small 7.6 & \small 24.8 & \small 42.2 & \small 22.7 & \small 0.0 & \small 24.3 & \small 2.9 & \small 30.5 & \small 10.1 \\

    \bottomrule
  \end{tabular}%
}
  \caption{Comparison of results for different models on the Argoverse 2 validation set.}
  \label{argvo}
\end{table*}

\begin{table}
  \centering
  \begin{tabular}{@{}lcccc@{}}
    \toprule
    Method & mAP & FPS & Params &Speedup \\
    \midrule
    SAFDNet\cite{safdnet} & 39.7 & 10.8&6.11M&1.0$\times$\\
     PointSlice-A (Ours)& 38.2&11.9 &4.05M&1.1$\times$ \\
    \bottomrule
  \end{tabular}
  \caption{Performance comparison of different models on the Argoverse 2 validation set.}
  \label{argojian}
\end{table}

\noindent\textbf{Results on Argoverse 2.} To further validate our model's detection performance in long-range and multi-class scenarios, we conducted experiments on the Argoverse 2 validation set. As shown in \cref{argvo} and \cref{argojian}, PointSlice achieves an mAP of 38.7, with a 1.0 mAP reduction compared to the state-of-the-art SAFDNet, but with a 1.10$\times$ faster inference speed and 0.66$\times$ the parameters. These results further demonstrate the efficiency of our model.

\begin{table}
  \centering
  \begin{tabular}{@{}lcccc@{}}
    \toprule
    Method & mAPH(L2) & Params & Speedup & FPS  \\
    \midrule
    SAFDNet   & 73.9   & 9.89M   & 1.0$\times$   & 13.68    \\
    SAFD-Pillar   & 69.3   & 7.74M  & 1.17$\times$ & 15.95  \\
    PointSlice-W & 72.7 & 7.82M & 1.13$\times$ & 15.40 \\
    \bottomrule
  \end{tabular}
  \caption{Ablation study of different point cloud processing methods on the Waymo validation set.}
  \label{ablation slice}
\end{table}

\subsection{Robustness Tests}

To thoroughly evaluate the reliability of our proposed PointSlice in real-world deployment scenarios, we conduct extensive robustness tests under diverse challenging conditions, including \textit{Distance Variation}, \textit{Point Cloud Sparsity}, and \textit{Coordinate Perturbation}. We compare our method against the high-performance baseline SAFDNet~\cite{safdnet}.

\noindent\textbf{Efficiency-Robustness Trade-off on Waymo.}
We extend our robustness analysis to the Waymo Open Dataset (~\cref{robusttests_all}), where our method maintains comparable stability to SAFDNet under varying degrees of sparsity and noise. Regarding distance robustness, as detailed in ~\cref{robusttests_distance}, PointSlice-W exhibits consistent accuracy across all ranges ($[0, 30)\text{m}$, $[30, 50)\text{m}$, and $[50, +\inf)\text{m}$). Even in the challenging long-range setting ($>50\text{m}$), the performance gap is negligible (e.g., 59.9\% vs. 60.4\% for Vehicle L1 AP), confirming that our lightweight design effectively preserves long-range geometric details. Crucially, these comparable results are achieved with significantly higher efficiency. As reported in ~\cref{waymovalidation}, PointSlice-W operates at \textbf{15.4 FPS} compared to SAFDNet's 13.6 FPS (a \textbf{13.2\%} speedup) and requires only \textbf{7.82M} parameters versus SAFDNet's 9.89M (a \textbf{20.9\%} reduction). Thus, our method offers a superior trade-off: delivering highly competitive accuracy and range robustness while being much more suitable for onboard deployment.

\noindent\textbf{Robustness against Sparsity on nuScenes.}
We further evaluate performance on the nuScenes validation set, where lidar sensors often face occlusion or distance-induced sparsity. We simulate this by randomly retaining a portion of points (ratios from $0.8$ to $0.3$). As detailed in ~\cref{nuscenes_robust}, our method demonstrates superior resilience compared to SAFDNet. Under the extreme sparsity setting (Ratio 0.3), PointSlice-N achieves an mAP of 56.5 and NDS of 64.6, surpassing SAFDNet by \textbf{1.6} in mAP and \textbf{0.8} in NDS. This indicates that our slice-based feature aggregation effectively preserves geometric structures even when point density is drastically reduced.

\noindent\textbf{Robustness against Sensor Noise on nuScenes.}
To simulate sensor noise, we apply Gaussian noise with standard deviations $\sigma \in \{0.05\text{m}, 0.10\text{m}\}$ to point coordinates. While both models are affected by coordinate perturbations, PointSlice-N exhibits better stability. Specifically, under severe noise ($\sigma=0.10\text{m}$), our method achieves 47.6 mAP, outperforming SAFDNet (43.7 mAP) by a margin of \textbf{3.9}. This suggests that our architecture is less sensitive to local coordinate jittering compared to the baseline.

\begin{table*}
  \centering 
\resizebox{\textwidth}{!}{%
  \begin{tabular}{@{}l|c|c|c|c|c|c@{}}
    \toprule
    \multirow{2}{*}{Method}  & \multicolumn{2}{c|}{ Vehicle AP/APH} & \multicolumn{2}{c|}{Pedestrian AP/APH} & \multicolumn{2}{c}{Cyclist AP/APH}  \\
                              & L1   & L2   & L1   & L2   & L1   & L2        \\
    \midrule
      \multicolumn{7}{c}{\textit{[0, 30) m}} \\
      \midrule
    
    SAFDNet \cite{safdnet} &  93.7/93.3 & 92.6/92.2 & 89.0/85.6 & 85.1/81.8 & 86.8/85.8 & 86.2/85.2 \\
    PointSlice-W & 93.5/93.1 & 92.4/92.1 & 88.1/84.5 & 83.6/80.0 & 86.1/85.1 & 85.4/84.5 \\
    \midrule
      \multicolumn{7}{c}{\textit{[30, 50) m}} \\
      \midrule
    
    SAFDNet \cite{safdnet} &  79.6/79.1 & 73.5/73.1 & 83.9/79.1 & 77.4/72.7 & 75.7/74.5 & 71.6/70.4 \\
    PointSlice-W & 79.2/78.7 & 73.1/72.6 & 82.9/77.6 & 76.2/71.2 & 74.1/73.0 & 70.1/69.1 \\
    \midrule
      \multicolumn{7}{c}{\textit{[50, +inf) m}} \\
      \midrule
    
    SAFDNet \cite{safdnet} &  60.4/59.7 & 48.0/47.4 & 77.1/69.7 & 64.2/57.5 & 64.0/63.0 & 59.7/58.7 \\
    PointSlice-W & 59.9/59.2 & 47.5/47.0 & 76.2/68.4 & 63.2/56.3 & 61.7/60.7 & 57.6/56.7 \\
    \bottomrule
  \end{tabular}%
  }
  \caption{Robustness analysis across different distance ranges on the Waymo validation set. We compare our PointSlice-W with the \textbf{high-performance} method SAFDNet~\cite{safdnet}. Both models are trained on the full training set (100\%).}
  \label{robusttests_distance}
\end{table*}

\begin{table}[htbp]
  \centering 
\resizebox{\textwidth}{!}{%
  \begin{tabular}{@{}l|c|c|c|c|c|c@{}}
    \toprule
    \multirow{2}{*}{Method}  & \multicolumn{2}{c|}{ Vehicle AP/APH} & \multicolumn{2}{c|}{Pedestrian AP/APH} & \multicolumn{2}{c}{Cyclist AP/APH}  \\
                              & L1   & L2   & L1   & L2   & L1   & L2        \\
    \midrule
      \multicolumn{7}{c}{\textit{Retention Ratio 1.0 (Original, No Noise)}} \\
      \midrule
      
    SAFDNet \cite{safdnet}$^{\dag}$&80.6/80.2&72.7/72.2&85.0/80.7&77.7/73.4&79.3/78.2&76.4/75.4\\
    PointSlice-W &80.3/79.8&72.3/71.9&84.1/79.4&76.4/71.9&77.9/76.9&75.1/74.2\\
    \midrule
      \multicolumn{7}{c}{\textit{Retention Ratio 0.8}} \\
      \midrule
      
    SAFDNet \cite{safdnet} &  79.9/79.4 & 71.8/71.4 & 84.0/79.1 & 76.3/71.6 & 76.7/75.6 & 73.8/72.9 \\
    PointSlice-W & 79.5/79.0 & 71.4/70.9 & 83.0/77.8 & 74.9/70.0 & 75.4/74.4 & 72.6/71.7 \\
    \midrule
      \multicolumn{7}{c}{\textit{Retention Ratio 0.6}} \\
      \midrule
      
    SAFDNet \cite{safdnet} &  78.6/78.1 & 70.3/69.9 & 81.8/76.3 & 73.7/68.5 & 72.1/71.1 & 69.4/68.5 \\
    PointSlice-W & 78.2/77.7 & 69.8/69.4 & 80.6/74.8 & 72.2/66.7 & 70.9/69.9 & 68.2/67.3 \\
    \midrule
      \multicolumn{7}{c}{\textit{Retention Ratio 0.5}} \\
      \midrule
      
    SAFDNet \cite{safdnet} &  77.5/77.0 & 69.1/68.7 & 79.9/74.1 & 71.6/66.1 & 68.3/67.3 & 65.8/64.8 \\
    PointSlice-W & 77.1/76.6 & 68.7/68.2 & 78.6/72.5 & 70.0/64.2 & 67.5/66.5 & 65.0/64.1 \\
    \midrule
      \multicolumn{7}{c}{\textit{Retention Ratio 0.3}} \\
      \midrule
      
    SAFDNet \cite{safdnet} &  73.2/72.8 & 64.7/64.3 & 72.8/65.7 & 64.2/57.7 & 54.5/53.6 & 52.4/51.6 \\
    PointSlice-W & 72.8/72.3 & 64.2/63.8 & 71.0/63.6 & 62.1/55.5 & 53.9/53.0 & 51.8/51.0 \\
    \midrule
      \multicolumn{7}{c}{\textit{Gaussian Noise ($\sigma=0.05$ m)}} \\
      \midrule
      
    SAFDNet \cite{safdnet} &  77.9/77.4 & 69.9/69.5 & 80.3/71.5 & 72.5/64.2 & 72.8/71.5 & 70.1/68.9 \\
    PointSlice-W & 77.5/77.0 & 69.5/69.1 & 79.4/70.5 & 71.2/62.9 & 72.1/70.9 & 69.4/68.3 \\
    \midrule
      \multicolumn{7}{c}{\textit{Gaussian Noise ($\sigma=0.10$ m)}} \\
      \midrule
      
    SAFDNet \cite{safdnet} &  33.7/33.1 & 29.6/29.1 & 29.3/20.2 & 25.3/17.4 & 29.1/27.3 & 28.0/26.2 \\
    PointSlice-W & 32.9/32.5 & 28.9/28.5 & 30.9/21.6 & 26.7/18.6 & 32.8/30.5 & 31.5/29.3 \\
    \bottomrule
  \end{tabular}%
  }
  \caption{Robustness comparison between SAFDNet and PointSlice on the Waymo validation set. We evaluate the models under two challenging conditions: point cloud sparsity (controlled by Retention Ratio) and coordinate perturbation (controlled by Gaussian Noise $\sigma$). Note that the sparsity tests and noise tests are conducted \textbf{independently} (i.e., noise tests use 100\% retention, and retention tests use zero noise).}
  \label{robusttests_all}
\end{table}

\begin{table*}
  \centering
  \resizebox{\textwidth}{!}{%
    \begin{tabular}{l|cc|cccccccccc}
      \hline
      Method & NDS & mAP  & Car & Truck  & Bus & T.L. & C.V. & Ped. & M.T. & Bike & T.C. & B.R. \\
      \midrule
      \multicolumn{13}{c}{\textit{Retention Ratio 1.0 (Original, No Noise)}} \\
      \midrule
      SAFDNet\cite{safdnet}$^{\dag}$ & 70.4 & 65.7 & 87.8 & 63.6 & 76.1 & 43.6 & 26.0 & 87.6 & 71.6 & 57.1 & 75.2 & 68.4 \\
      PointSlice-N (Ours) & 70.9 & \textbf{66.7} & 87.9 & 63.8 & 78.3 & 47.8 & 26.0 & 88.1 & 73.7 & 58.0 & 75.8 & 68.1 \\
      \midrule
      \multicolumn{13}{c}{\textit{Retention Ratio 0.8}} \\
      \midrule
      SAFDNet\cite{safdnet} & 69.6 & 64.6 & 86.4 & 62.8 & 76.0 & 43.2 & 25.5 & 86.7 & 68.9 & 55.1 & 73.1 & 68.1 \\
      PointSlice-N (Ours) & 70.3 & 65.7 & 86.4 & 63.0 & 77.7 & 47.1 & 25.3 & 87.2 & 73.0 & 56.1 & 73.8 & 67.6 \\
      \midrule
      \multicolumn{13}{c}{\textit{Retention Ratio 0.6}} \\
      \midrule
      SAFDNet\cite{safdnet} & 68.4 & 62.6 & 84.2 & 61.5 & 75.4 & 42.7 & 24.4 & 84.4 & 64.5 & 51.8 & 69.6 & 67.4 \\
      PointSlice-N (Ours) & 69.1 & 63.7 & 84.0 & 61.9 & 77.0 & 46.3 & 24.6 & 84.8 & 69.3 & 52.7 & 70.0 & 66.7 \\
      \midrule
      \multicolumn{13}{c}{\textit{Retention Ratio 0.5}} \\
      \midrule
      SAFDNet\cite{safdnet} & 67.5 & 60.9 & 82.6 & 60.4 & 74.8 & 41.9 & 23.6 & 82.5 & 61.4 & 48.7 & 66.7 & 66.9 \\
      PointSlice-N (Ours) & 68.1 & 62.2 & 82.4 & 60.8 & 76.5 & 45.7 & 23.8 & 83.0 & 66.7 & 49.9 & 67.1 & 66.1 \\
      \midrule
      \multicolumn{13}{c}{\textit{Retention Ratio 0.3}} \\
      \midrule
      SAFDNet\cite{safdnet} & 63.8 & 54.9 & 76.7 & 55.8 & 71.2 & 40.2 & 20.4 & 74.1 & 50.2 & 38.8 & 57.7 & 64.2 \\
      PointSlice-N (Ours) & 64.6 & 56.5 & 76.3 & 56.6 & 73.4 & 43.6 & 21.2 & 75.0 & 56.8 & 40.8 & 58.1 & 63.4 \\
      \midrule
      \multicolumn{13}{c}{\textit{Gaussian Noise ($\sigma=0.05$ m)}} \\
      \midrule
      SAFDNet\cite{safdnet} & 68.2 & 62.3 & 86.6 & 62.0 & 72.8 & 41.2 & 20.8 & 83.9 & 66.3 & 50.8 & 71.2 & 67.6 \\
      PointSlice-N (Ours) & 68.8 & 63.6 & 86.6 & 61.7 & 76.2 & 46.5 & 20.9 & 83.9 & 68.6 & 53.6 & 70.1 & 67.8 \\
      \midrule
      \multicolumn{13}{c}{\textit{Gaussian Noise ($\sigma=0.1$ m)}} \\
      \midrule
      SAFDNet\cite{safdnet} & 56.9 & 43.7 & 79.9 & 44.8 & 51.0 & 33.0 & 6.3 & 65.5 & 45.1 & 21.3 & 49.8 & 39.9 \\
      PointSlice-N (Ours) & 59.0 & 47.6 & 81.2 & 46.8 & 61.4 & 34.6 & 7.8 & 67.4 & 50.9 & 27.1 & 52.2 & 46.4 \\
      \midrule
      
    \end{tabular}%
  }
  \caption{Robustness results on the nuScenes validation set. $^{\dag}$ denotes results reproduced by us. ‘T.L.’, ‘C.V.’, ‘Ped.’, ‘M.T.’, ‘T.C.’, and ’B.R.’ denote trailer, construction vehicle, pedestrian, motor, traffic cone, and barrier.}
  \label{nuscenes_robust}
\end{table*}

\subsection{Ablation studies}
To validate the effectiveness of the PointSlice model over the pillar-based approach in point cloud processing, as well as the contribution of the SIN module, we conducted ablation experiments on the Waymo dataset.

\noindent
\textbf{Comparison between slice and pillar approaches.} To ensure variable consistency, our comparison isolates the difference in point cloud processing methods. We keep the 2D feature map dimensions, the number of dimensions, and the quantity of SRB and EDB modules consistent between both methods. SAFDNet-pillar denotes a configuration where each voxel size is set to $[0.08, 0.08, 6]$, allowing us to observe the performance of SAFDNet under pillar-based processing. As shown in \cref{ablation slice}, PointSlice achieves a higher accuracy of +3.4\% mAP compared to SAFDNet-pillar while maintaining similar inference speed. This demonstrates that the slice approach provides more precise results than pillar processing for point clouds, while also preserving model efficiency. 
Consequently, this capability to deliver superior perception fidelity without increasing latency makes PointSlice highly practical for real-world autonomous driving systems. Specifically, the balanced efficiency ensures synchronization with standard 15Hz sensors and alleviates memory constraints, providing reliable support for critical downstream tasks such as path planning and obstacle avoidance.

\begin{table}
  \centering
  \begin{tabular}{@{}lcc@{}}
    \toprule
    Method & mAPH(L2) & Params  \\
    \midrule
    (2D-STEM,2D-EDB)   & 71.5  & 7.62M      \\
    (2D-STEM+SIN,2D-EDB)   & 72.5   & 7.71M   \\
    (2D-STEM,2D-EDB+SIN)   & 71.9   & 7.73M   \\
    (2D-STEM+SIN,2D-EDB+SIN) & 72.7 & 7.82M  \\
    \bottomrule
  \end{tabular}
  \caption{Ablation study of applying SIN at different positions in PointSlice on the Waymo validation set.}
  \label{SINablation}
\end{table}

\noindent
\textbf{Comparison of SIN Modules.} To demonstrate the role of SIN in PointSlice, we conducted three comparative experiments, as shown in \cref{SINablation}. The first row indicates that the SIN module is not included in the PointSlice network, while the second row represents the insertion of the SIN module only in the 2D-STEM. A comparison of these two results shows that adding the SIN module improves the model's accuracy by 1.0\%, indicating that SIN indeed facilitates feature interaction across slices. The third row indicates that adding the SIN module solely within the 2D-EDB module results in a 0.4 improvement in performance compared to the first row without SIN. The fourth row represents the model with the SIN module integrated into both the 2D-STEM and 2D-EDB modules, achieving the best performance with an mAPH of 72.7. This demonstrates that the SIN module plays a crucial role in enhancing both the 2D-STEM and 2D-EDB modules.

\noindent
\textbf{Comparison of Memory Usage During Model Inference.} We conducted experiments to evaluate GPU memory consumption of different models on various datasets. To measure memory usage, we utilized torch.cuda.max\_memory\_allocated().
\begin{table}
  \centering
  \begin{tabular}{lccc}
    \hline
    \multirow{2}{*}{Model} & \multicolumn{3}{c}{Memory-Inference} \\  
                            & Waymo & nuScenes & Argoverse 2  \\ \hline
    SAFDNet\cite{safdnet}                & 410MB   & 905MB   & 273MB\\ 
     
    PointSlice               & 264MB   & 860MB 
    & 188MB 
    \\
    \hline
  \end{tabular}
  \caption{Comparison of Model GPU Memory Usage on Different Datasets. For all experiments, the batch size is set to 1.}
  \label{memory}
\end{table}

The experimental results are shown in \cref{memory}. From the table, it can be observed that on the Waymo, nuScenes, and Argoverse 2 datasets, the memory consumption of our PointSlice model is only 0.64 times, 0.95 times, and 0.69 times that of the SAFDNet model, respectively. This demonstrates the efficiency and lightweight characteristics of our proposed model.

\noindent
\textbf{Sensitivity Analysis of Slice Granularity.} The vertical resolution of the slice representation is a critical hyperparameter governing the trade-off between feature granularity and computational efficiency. To investigate this, we perform a sensitivity analysis on the slice height $K$ (and the corresponding slice number $H$) using the Waymo validation set. Specifically, the slice number is determined by $H = Z_{range} / K$, where $Z_{range}$ represents the total vertical range of the point cloud. As shown in ~\cref{waymochoose}, finer slices (e.g., $K=0.05$ m) preserve more geometric details, yielding the highest accuracy for the \textit{Cyclist} class (73.6\% L1 AP), but at the cost of increased computational burden and slightly lower inference speed (15.6 FPS). In contrast, increasing the slice height to $K=0.15$ m improves the performance for both \textit{Vehicle} and \textit{Pedestrian} classes, reaching peak AP scores of 77.2\% and 81.1\%, respectively. This suggests that $H=41$ provides sufficient vertical context while suppressing noise or redundant information often present in overly fine-grained partitions. While coarser slices ($K \ge 0.20$ m) lead to higher inference speeds (up to 16.6 FPS), they suffer from a noticeable degradation in detection accuracy due to the loss of fine-grained spatial information. Consequently, we select $K=0.15$ m as the default setting for our final model to achieve the optimal balance between accuracy and speed.
\begin{table*}
  \centering
  \resizebox{\textwidth}{!}{%
  \begin{tabular}{@{}c|c|c|c|c|c|c|c|c@{}}
    \toprule
    \multirow{2}{*}{Method} & \multirow{2}{*}{Height (m)/Number} & \multicolumn{2}{c|}{Vehicle AP/APH} & \multicolumn{2}{c|}{Pedestrian AP/APH} & \multicolumn{2}{c|}{Cyclist AP/APH} & \multirow{2}{*}{FPS} \\
     & & L1 & L2 & L1 & L2 & L1 & L2 & \\
    \midrule
    
    \multirow{5}{*}{\rotatebox[origin=c]{90}{PointSlice-W}} 
      & 0.05 / 121 & 76.3/75.8 & 67.9/67.5 & 80.4/74.8 & 72.3/67.0 & \textbf{73.6/72.5} & \textbf{70.9/69.9} & 15.6 \\
    
      & 0.10 / 61 & 76.8/76.3 & 68.4/68.0 & 80.7/74.8 & 72.7/67.2 & 73.1/72.0 & 70.4/69.4 & 16.1 \\
      
      & 0.15 / 41 & \textbf{77.2/76.7} & \textbf{69.0/68.6} & \textbf{81.1/75.1} & \textbf{73.2/67.5} & 73.1/72.0 & 70.4/69.4 & 16.4 \\
      
      & 0.20 / 31 & 77.0/76.5 & 68.8/68.3 & 81.0/75.0 & 73.0/67.3 & 72.8/71.8 & 70.1/69.1 & \textbf{16.6} \\
      
      & 0.25 / 25 & 76.9/76.4 & 68.6/68.2 & 80.7/74.3 & 72.7/66.8 & 72.5/71.4 & 69.8/68.8 & \textbf{16.6} \\
    
    \bottomrule
  \end{tabular}%
  }
  \caption{Sensitivity analysis of the slice number on the Waymo validation set using 20\% of the training data. The column ``Height/Number'' denotes the slice height and the corresponding total number of slices, respectively. Bold values indicate the best performance.}
  \label{waymochoose}
\end{table*}

\noindent
\textbf{Scalability Analysis.} 
To investigate the scalability of our proposed framework, we conducted a series of experiments by increasing the model depth, as summarized in ~\cref{scalabity testing}. Specifically, we enhanced the feature extraction capacity by inserting additional 3D Sparse Residual Blocks (SRBs) prior to the Slice Interaction Network (SIN). The architecture of the employed 3D SRB is analogous to the 2D block illustrated in ~\cref{fig:2dsrb}, with the modification of replacing 2D Submanifold Sparse Convolutions with their 3D counterparts to facilitate multi-slice interaction in 3D space.

Quantitative results demonstrate that increasing the number of interaction layers yields consistent performance gains. As shown in the 100\% data setting, the deeper variants successfully narrow the performance gap with the heavy baseline SAFDNet, particularly for smaller objects. For instance, PointSlice-W-D2 achieves 73.2\% and 75.1\% in L2 APH for Pedestrians and Cyclists, respectively. While this improvement incurs a slight decrease in inference speed (from 15.4 FPS to 14.4 FPS), the model maintains a competitive balance between accuracy and efficiency. These results confirm that PointSlice is highly scalable and can be effectively extended by stacking slice interaction modules to meet higher accuracy requirements.

\begin{table*}
  \centering
  \resizebox{\textwidth}{!}{%
  \begin{tabular}{@{}l|c|c|c|c|c|c|c|c@{}}
    \toprule
    \multirow{2}{*}{Method}  & \multicolumn{2}{c|}{ Vehicle AP/APH} & \multicolumn{2}{c|}{Pedestrian AP/APH} & \multicolumn{2}{c|}{Cyclist AP/APH} & \multirow{2}{*}{FPS} &\multirow{2}{*}{Params} \\
                              & L1   & L2   & L1   & L2   & L1   & L2   &      \\
    
    \midrule
      \multicolumn{9}{c}{\textit{Training with 20\% data}} \\
      \midrule
    SAFDNet\cite{safdnet}&79.1/78.6&71.0/70.5&83.6/78.7&75.9/71.2&77.8/76.6&74.9/73.8&13.6&9.89M\\
    PointSlice-W &78.8/78.3&70.7/70.3&82.9/77.4&75.3/70.0&75.4/74.3&72.6/71.6&15.4&7.82M\\

    PointSlice-W-D  &  78.7/78.2 & 70.5/70.1 & 83.1/78.0 & 75.4/70.5 & 76.4/75.3 & 73.6/72.5 &15.1 &8.11M\\

    \midrule
      \multicolumn{9}{c}{\textit{Training with 100\% data}} \\
      \midrule
    SAFDNet\cite{safdnet}&80.6/80.1&72.7/72.3&84.7/80.4&77.3/73.1&80.0/79.0&77.2/76.2&13.6&9.89M\\
    PointSlice-W &80.3/79.8&72.3/71.9&84.1/79.4&76.4/71.9&77.9/76.9&75.1/74.2&15.4&7.82M\\
    PointSlice-W-D &80.4/79.9&72.4/72.0&84.2/79.8&76.8/72.5&79.4/78.4&76.5/75.6&15.1&8.11M\\
    PointSlice-W-D2 &80.5/80.0&72.5/72.1&84.9/80.6&77.3/73.2&78.9/77.9&76.0/75.1&14.4&8.40M\\

    \bottomrule
  \end{tabular}%
  }
  
  \caption{Scalability analysis of PointSlice on the Waymo Open Dataset. We evaluate the effect of increasing the network depth by adding 3D Sparse Residual Blocks (SRBs). \textbf{PointSlice-W-D} and \textbf{PointSlice-W-D2} denote the variants with one and two additional SRBs inserted before the SIN module, respectively.}
  \label{scalabity testing}
\end{table*}

\subsection{Visualization of Feature Interaction}
\label{sec:visualization}

To intuitively investigate the inter-slice interaction mechanism facilitated by the SIN module, we conduct a qualitative analysis on the Waymo Open Dataset. Specifically, we visualize and compare the Bird's-Eye-View (BEV) feature maps before and after passing through the stacked SIN modules (4 stages). To generate the visualization, we aggregate the high-dimensional 3D feature volumes into 2D heatmaps by computing the mean value along both the height and channel dimensions. Subsequently, min-max normalization is applied to map the feature magnitudes to the range $[0, 1]$ for better visual contrast.

\begin{figure*}[htbp] 
    \centering
    

    \begin{subfigure}[b]{0.48\textwidth}
        \centering
        \includegraphics[width=\textwidth]{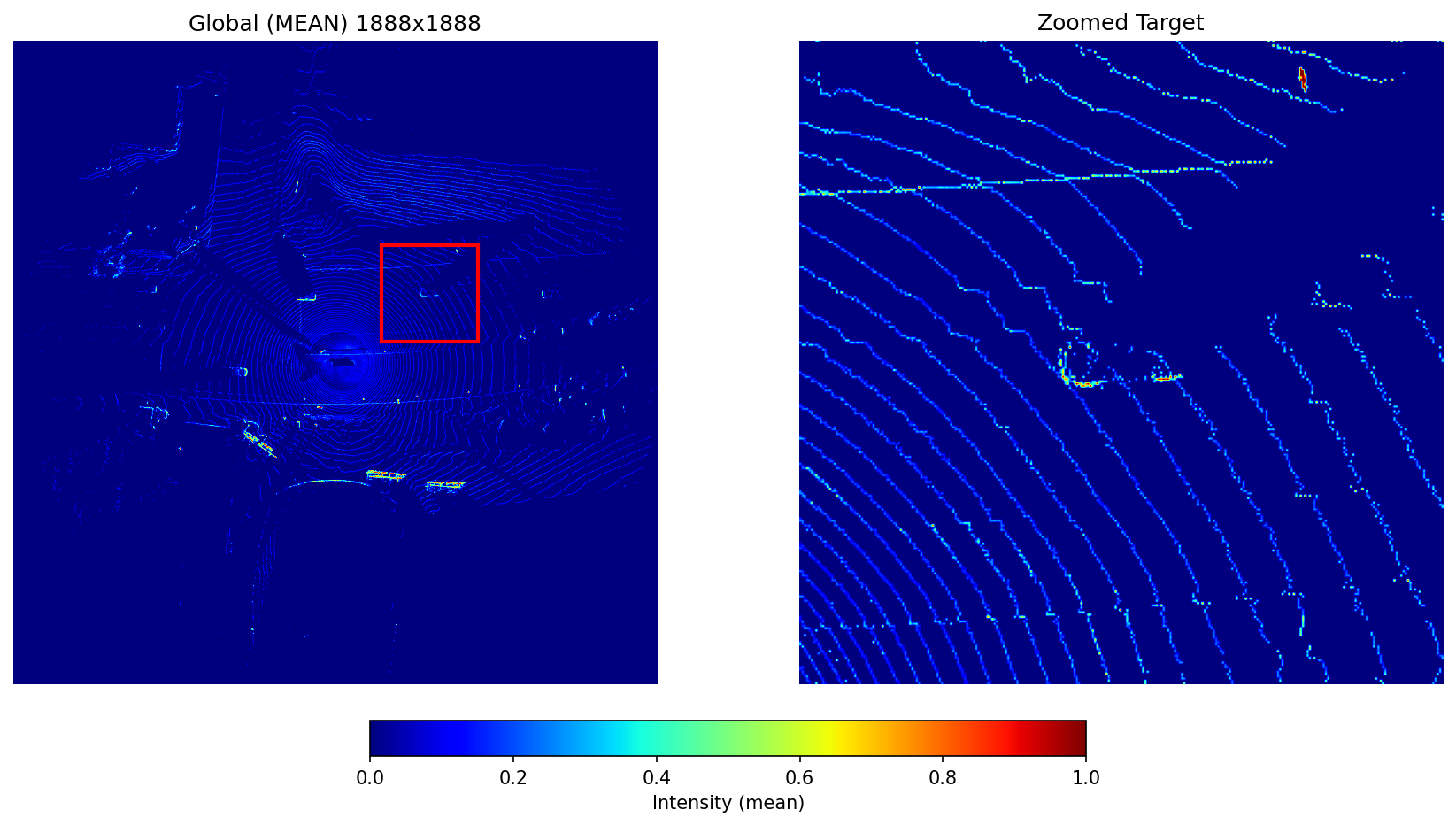}
        \caption{Stage 1: Before SIN}
        \label{fig:stage1_before}
    \end{subfigure}
    \hfill
    \begin{subfigure}[b]{0.48\textwidth}
        \centering
        \includegraphics[width=\textwidth]{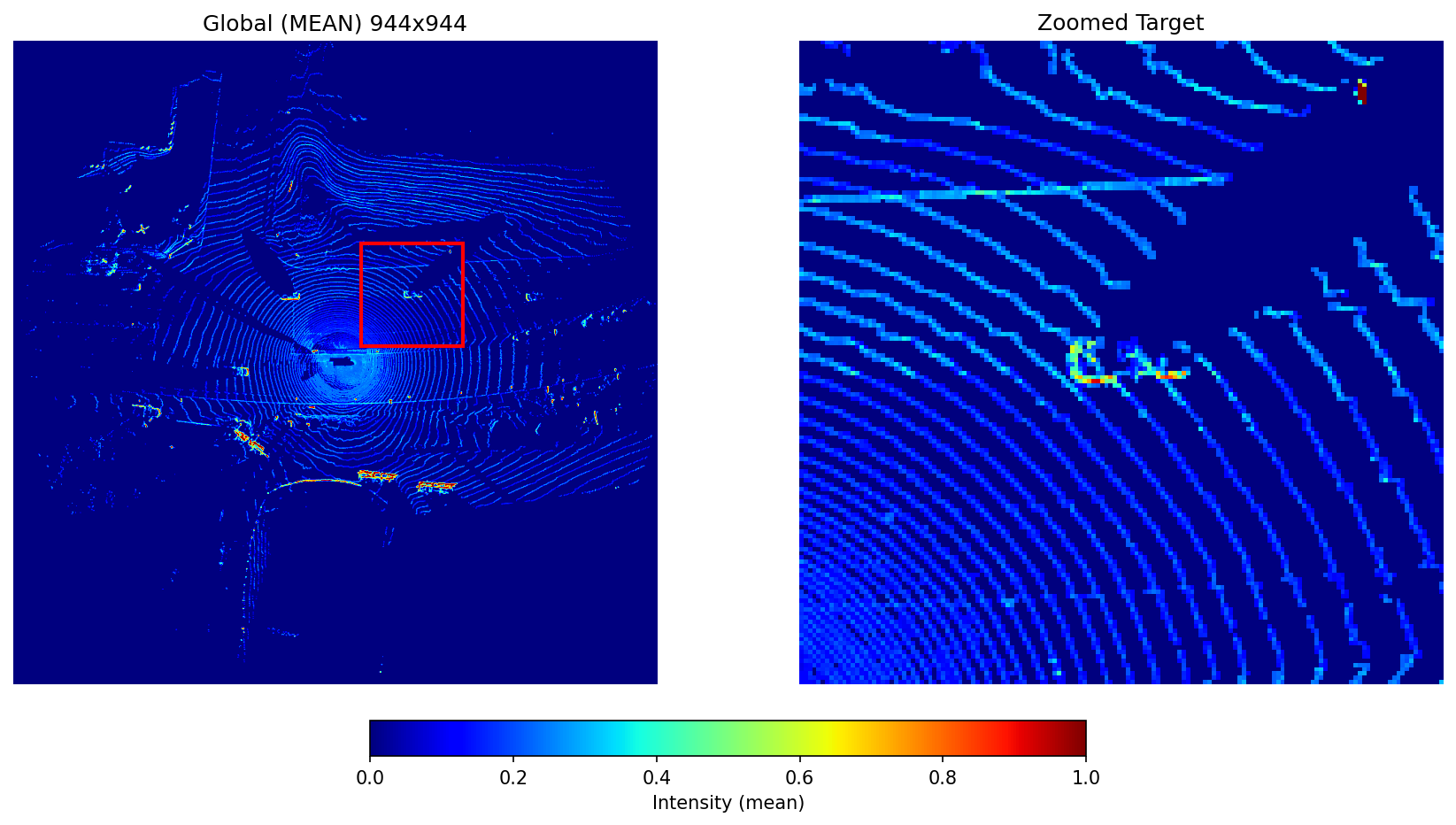}
        \caption{Stage 1: After SIN}
        \label{fig:stage1_after}
    \end{subfigure}

    \vspace{0.5em}

    \begin{subfigure}[b]{0.48\textwidth}
        \centering
        \includegraphics[width=\textwidth]{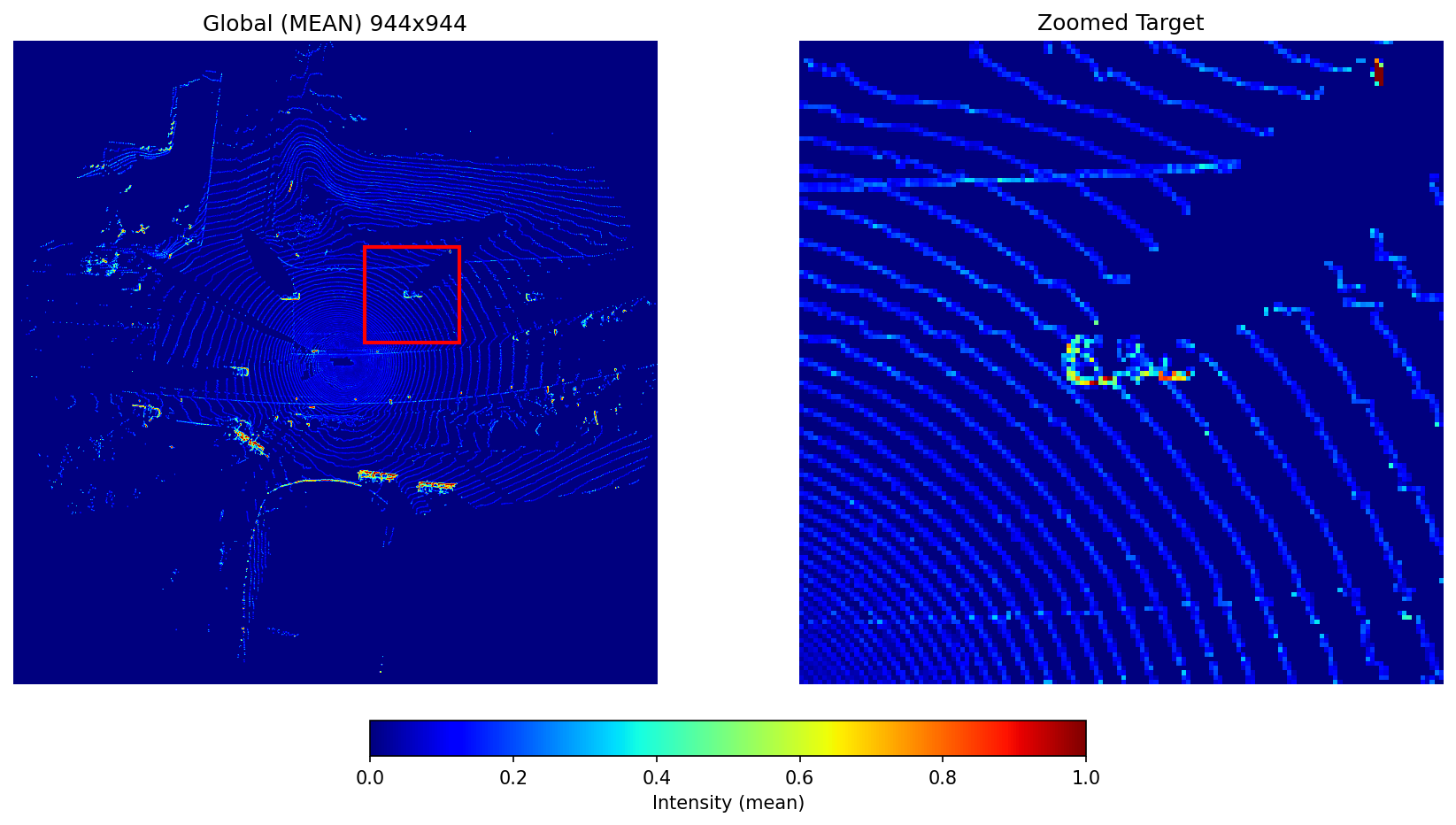}
        \caption{Stage 2: Before SIN}
        \label{fig:stage2_before}
    \end{subfigure}
    \hfill
    \begin{subfigure}[b]{0.48\textwidth}
        \centering
        \includegraphics[width=\textwidth]{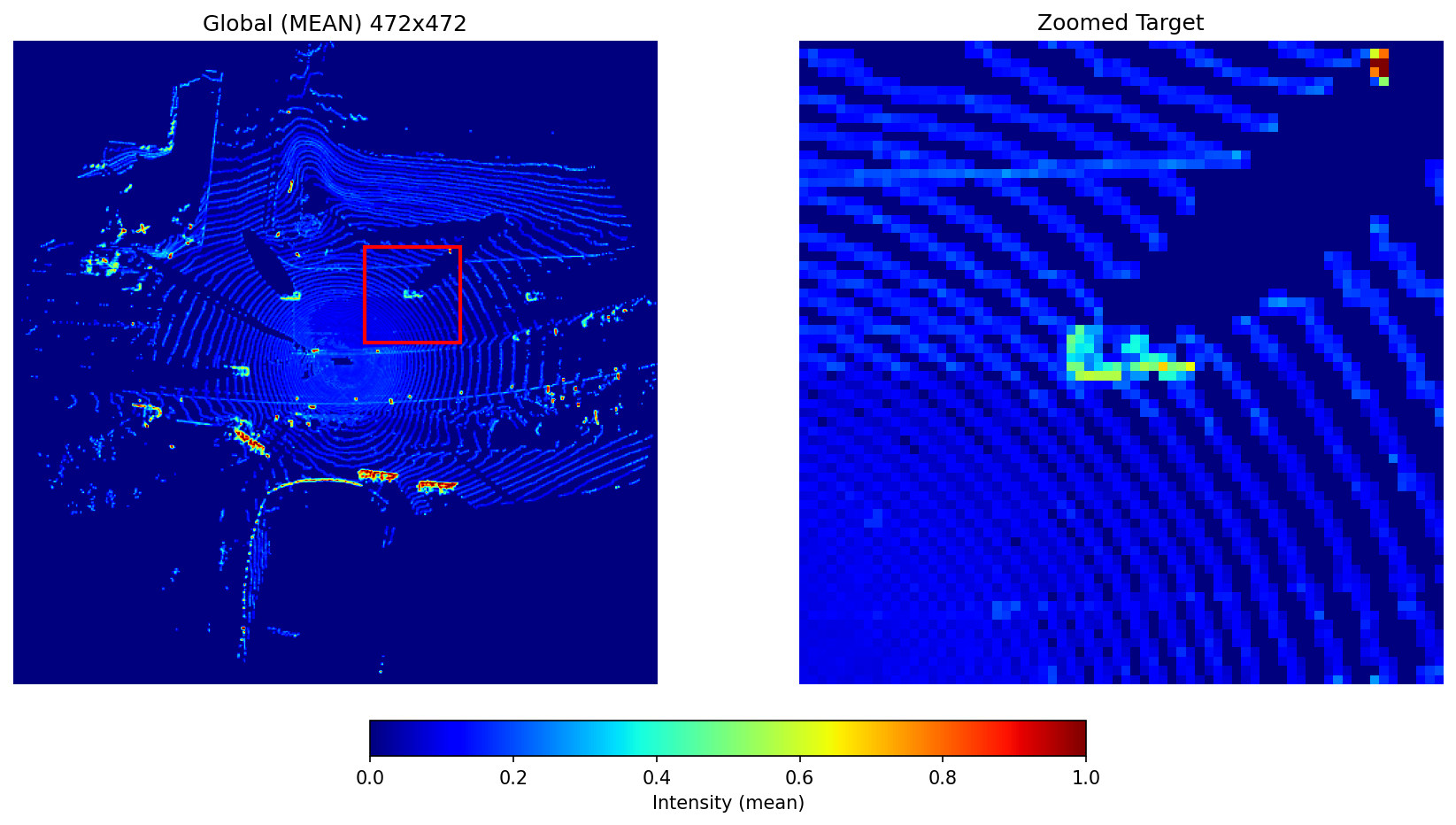}
        \caption{Stage 2: After SIN}
        \label{fig:stage2_after}
    \end{subfigure}

    \vspace{0.5em}

    \begin{subfigure}[b]{0.48\textwidth}
        \centering
        \includegraphics[width=\textwidth]{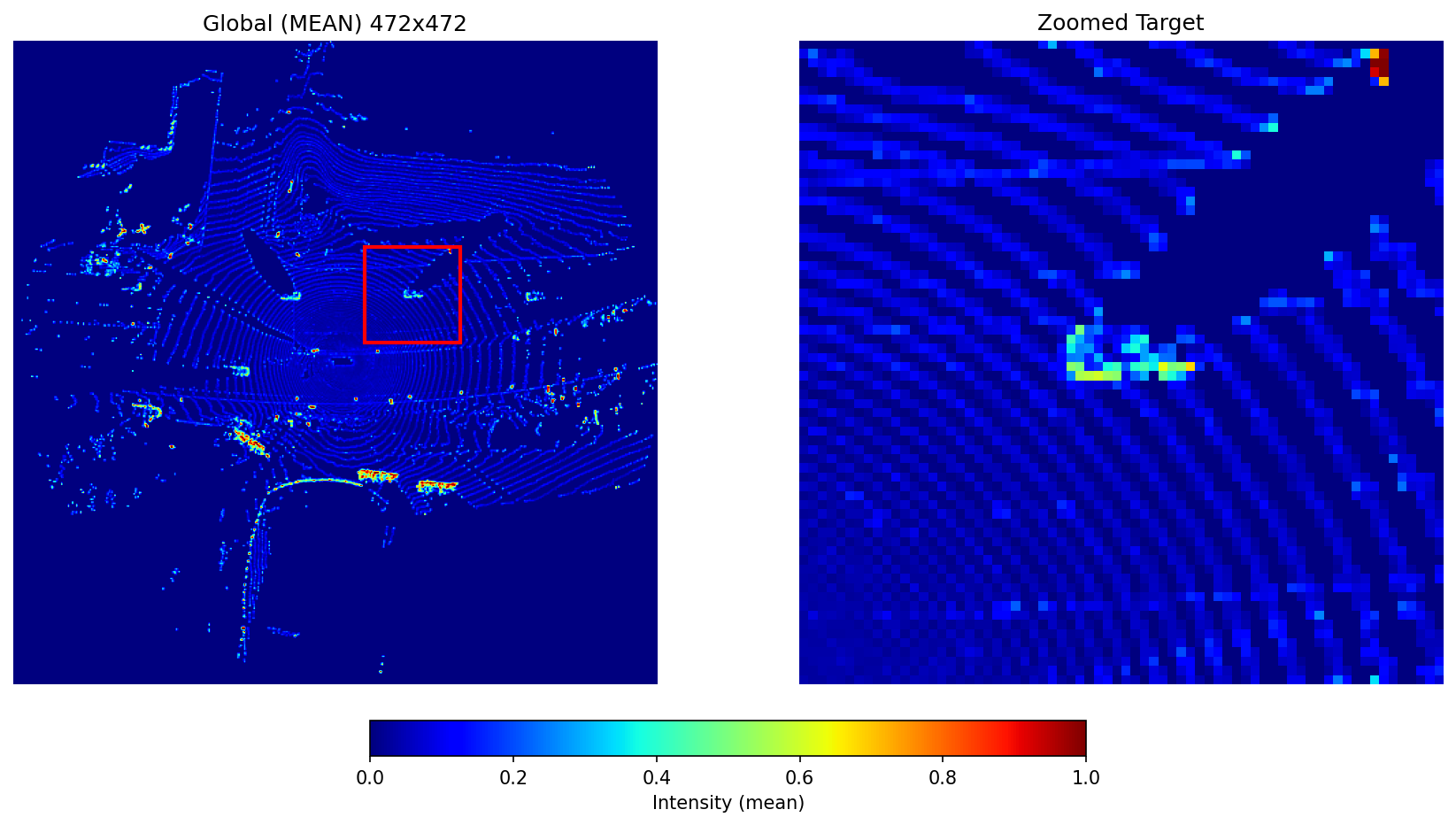}
        \caption{Stage 3: Before SIN}
        \label{fig:stage3_before}
    \end{subfigure}
    \hfill
    \begin{subfigure}[b]{0.48\textwidth}
        \centering
        \includegraphics[width=\textwidth]{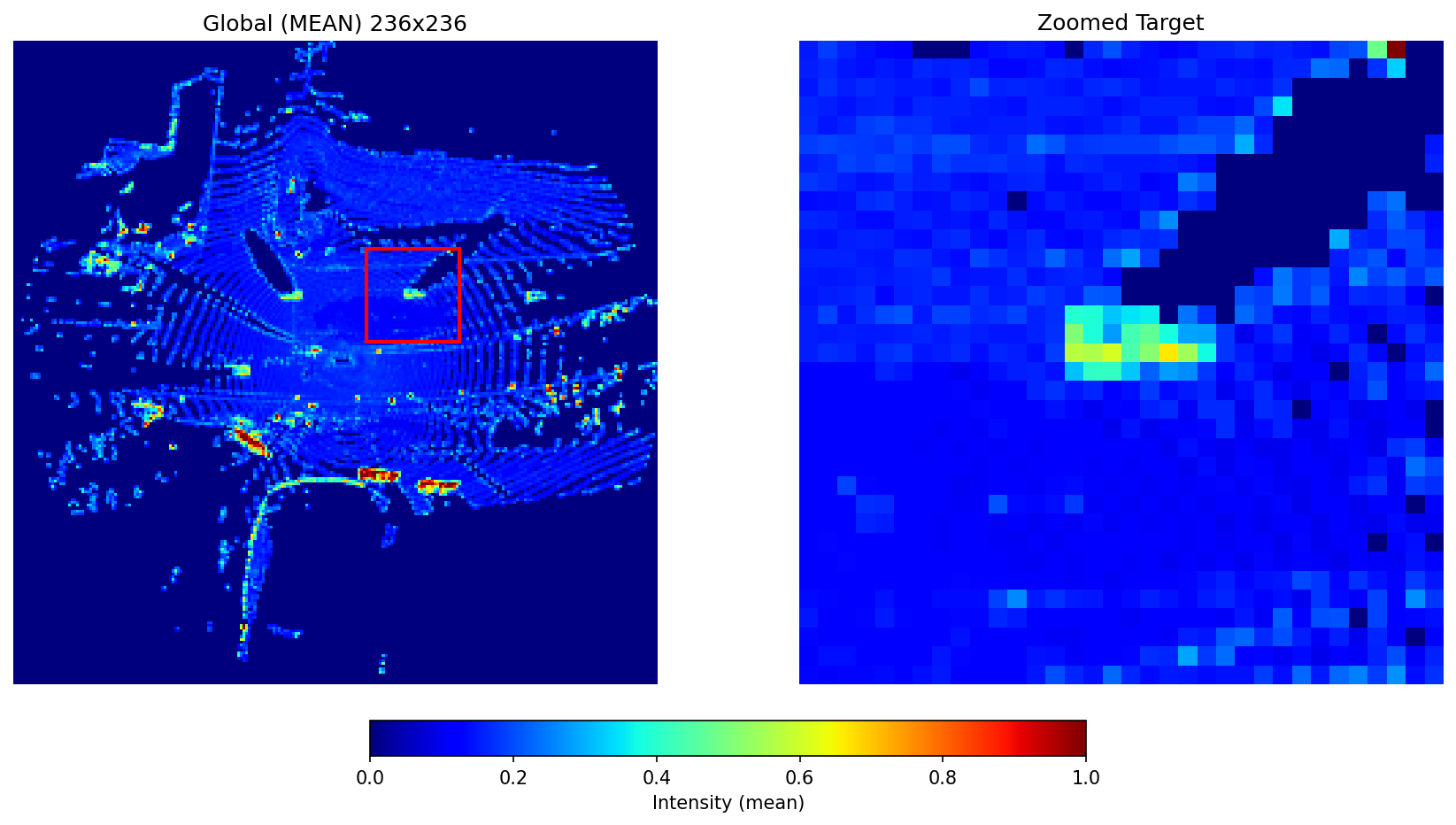}
        \caption{Stage 3: After SIN}
        \label{fig:stage3_after}
    \end{subfigure}
    
\vspace{0.5em} 
    \begin{subfigure}[b]{0.48\textwidth}
        \centering
        \includegraphics[width=\textwidth]{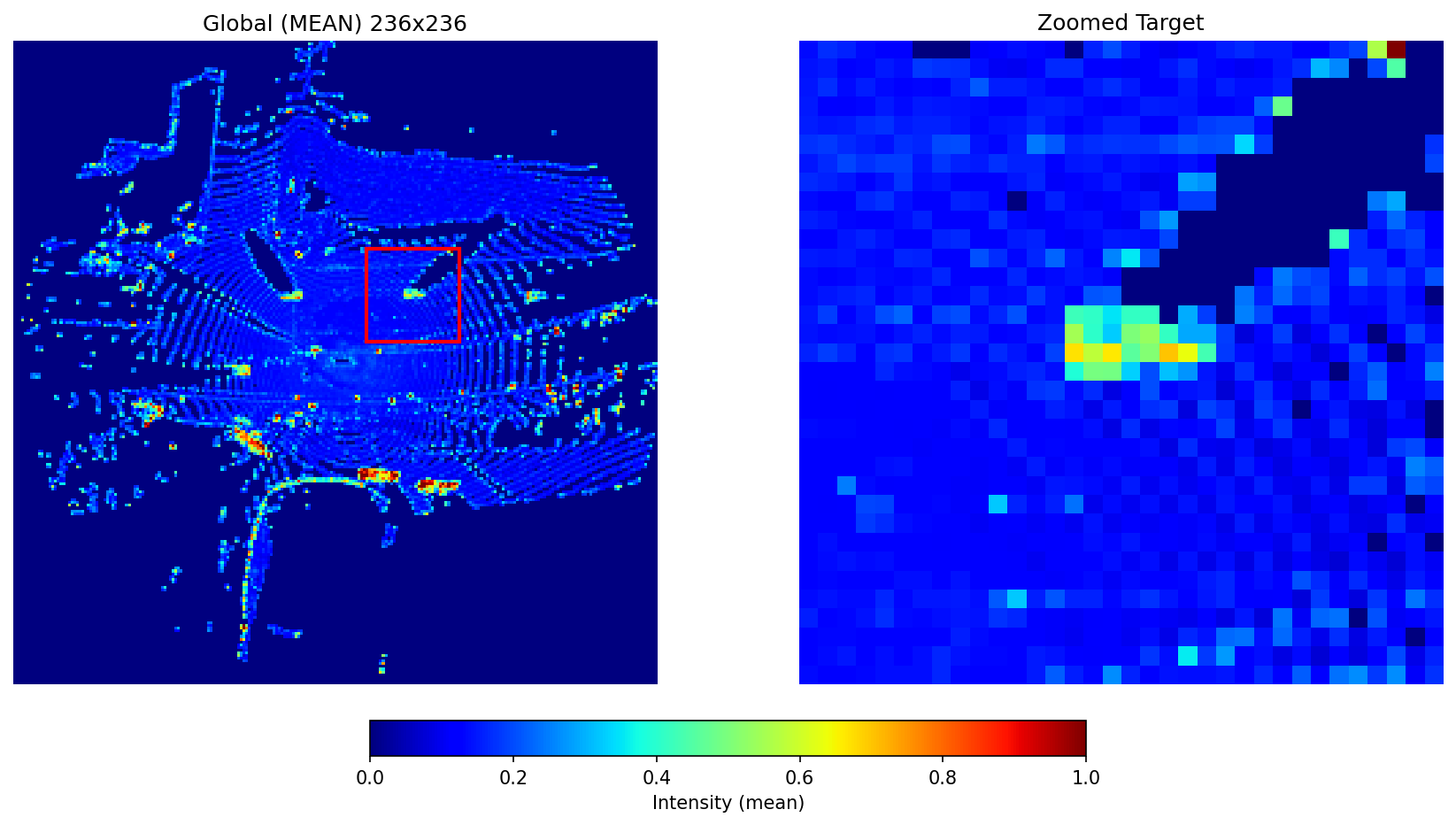}
        \caption{Before EDB\_SIN}
        \label{fig:before_sedsin}
    \end{subfigure}
    \hfill
    \begin{subfigure}[b]{0.48\textwidth}
        \centering
        \includegraphics[width=\textwidth]{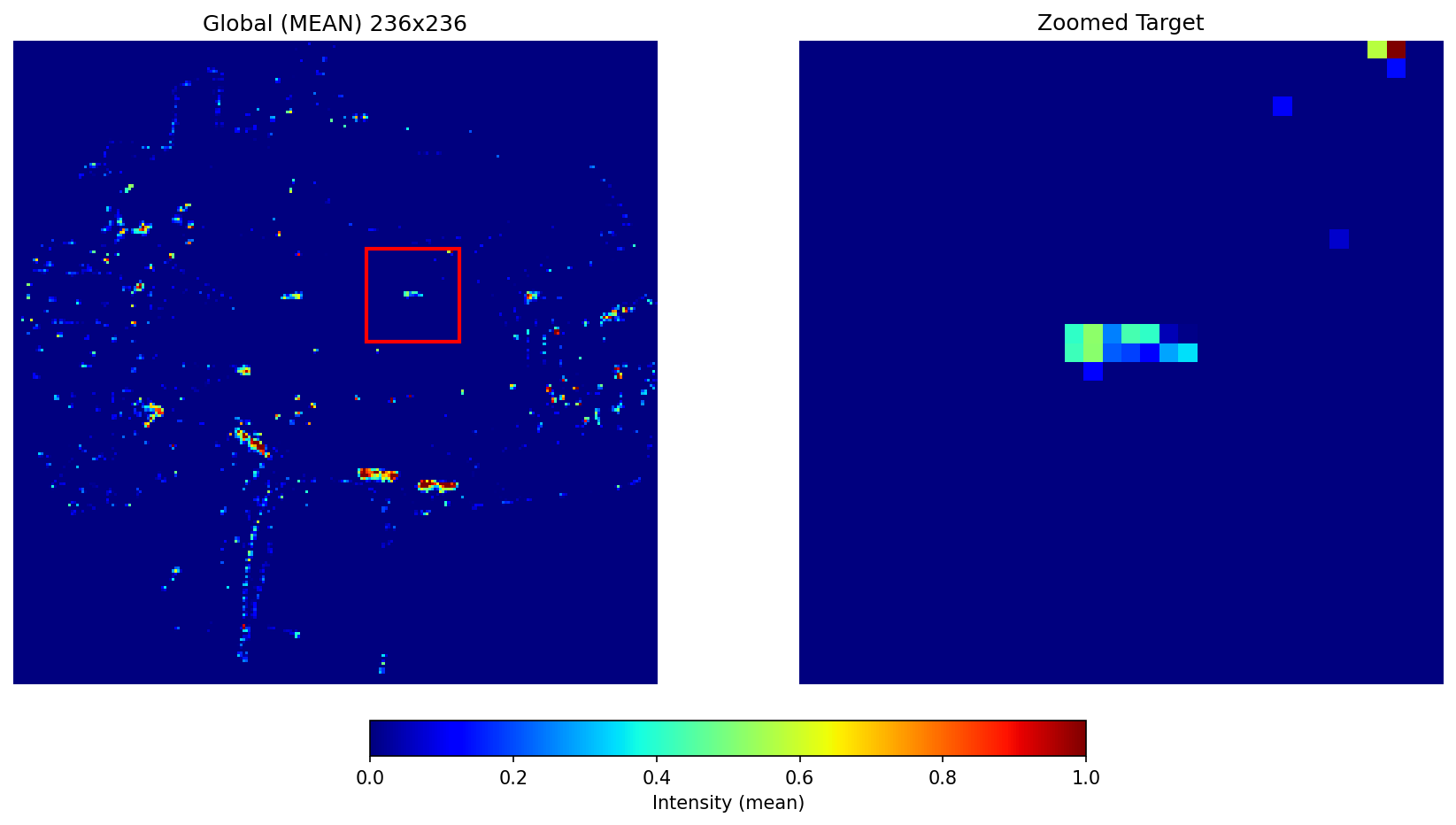}
        \caption{After EDB\_SIN}
        \label{fig:after_sedsin}
    \end{subfigure}

    \caption{Visual comparison of feature maps before and after processing at different stages. The left column shows the features before the SIN module, and the right column shows the enhanced features after interaction.}
    \label{fig:vis_feature_maps}
\end{figure*}

As illustrated in \cref{fig:vis_feature_maps}, the comparison demonstrates the effectiveness of the slice interaction in two main aspects. First, regarding the foreground-background contrast, the feature interaction effectively suppresses background noise after SIN processing. Consequently, the foreground objects (i.e., detection targets) exhibit significantly stronger activation responses relative to the background compared to the initial features. Second, we observe a clear feature densification effect, where the voxel representations within the foreground object regions become noticeably denser and more complete. This suggests that the SIN module effectively propagates semantic information across slices, filling in sparse gaps within the object geometry and resulting in more robust feature representations.

\subsection{Qualitative visualization}
\begin{figure*}[htbp]
    \centering
    \begin{subfigure}[t]{0.47\textwidth}
        \centering
        \includegraphics[height=5cm,width=\textwidth]{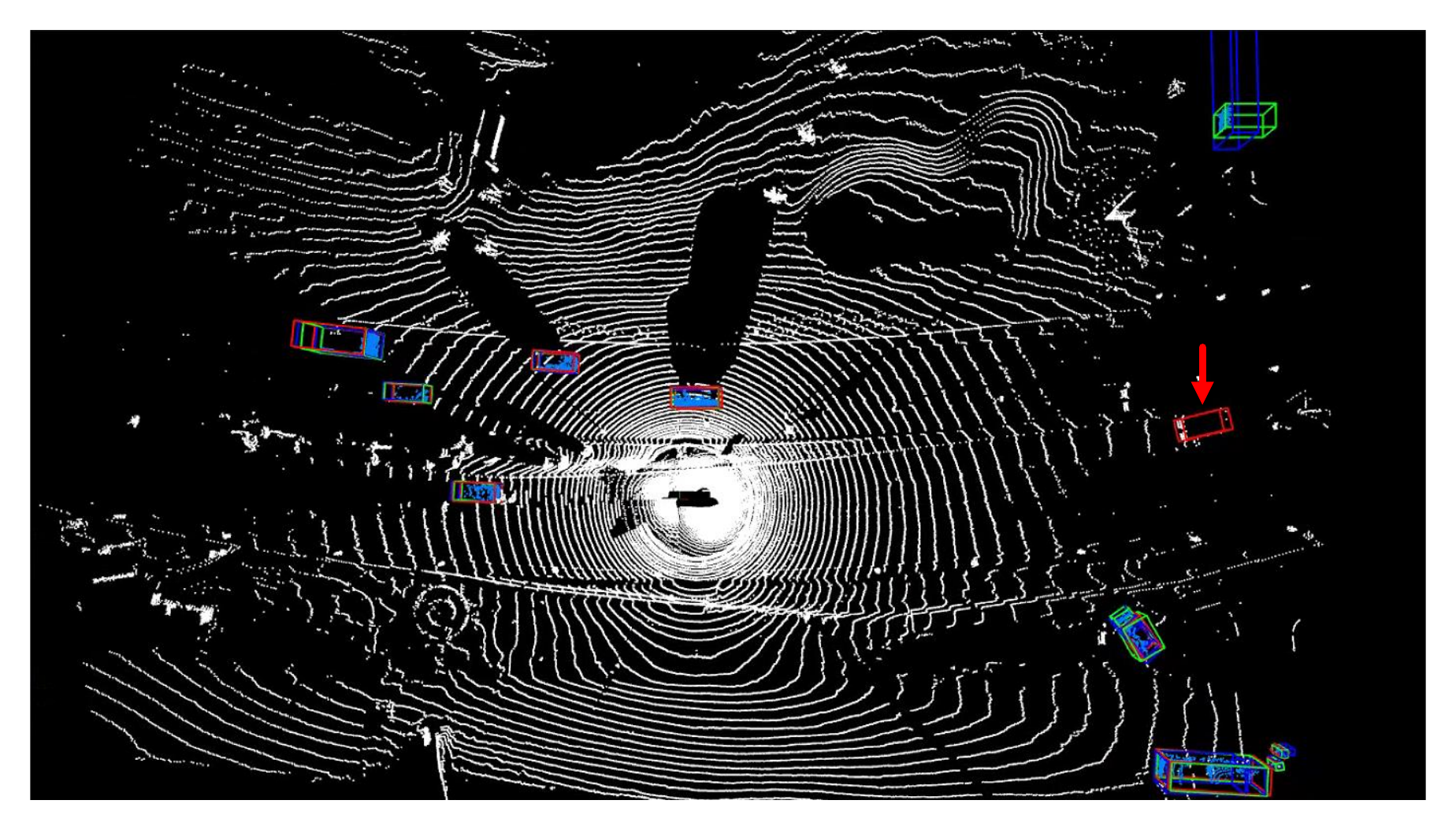} 
        \caption{Image(1)}
        \label{fig:image1}
    \end{subfigure}
    \hspace{0.02\textwidth}  
    \begin{subfigure}[t]{0.47\textwidth}
        \centering
        \includegraphics[height=5cm,width=\textwidth]{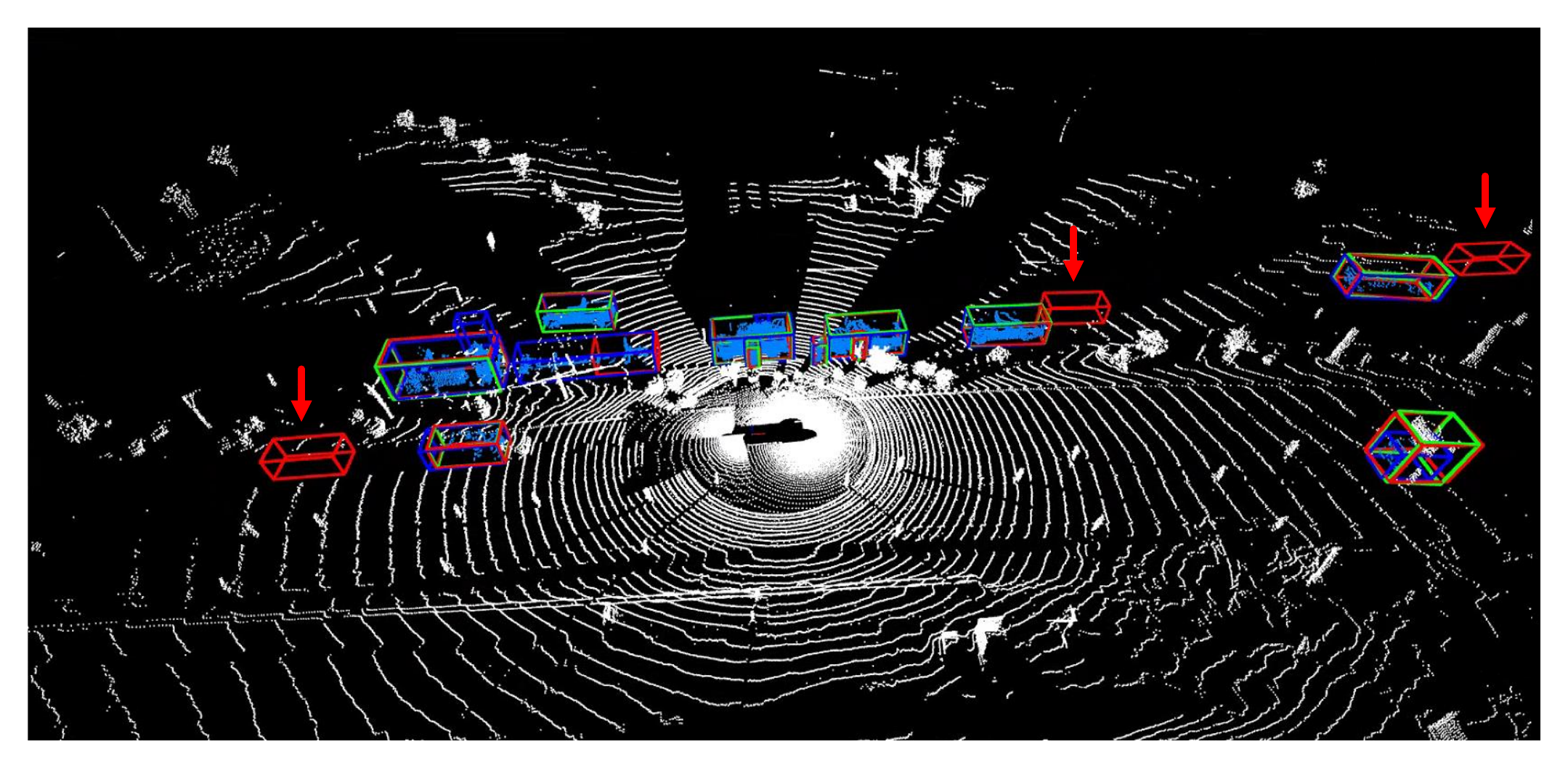} 
        \caption{Image(136)}
        \label{fig:image2}
    \end{subfigure}
    \vspace{0.5cm}  
    \begin{subfigure}[t]{0.47\textwidth}
        \centering
        \includegraphics[height=5cm,width=\textwidth]{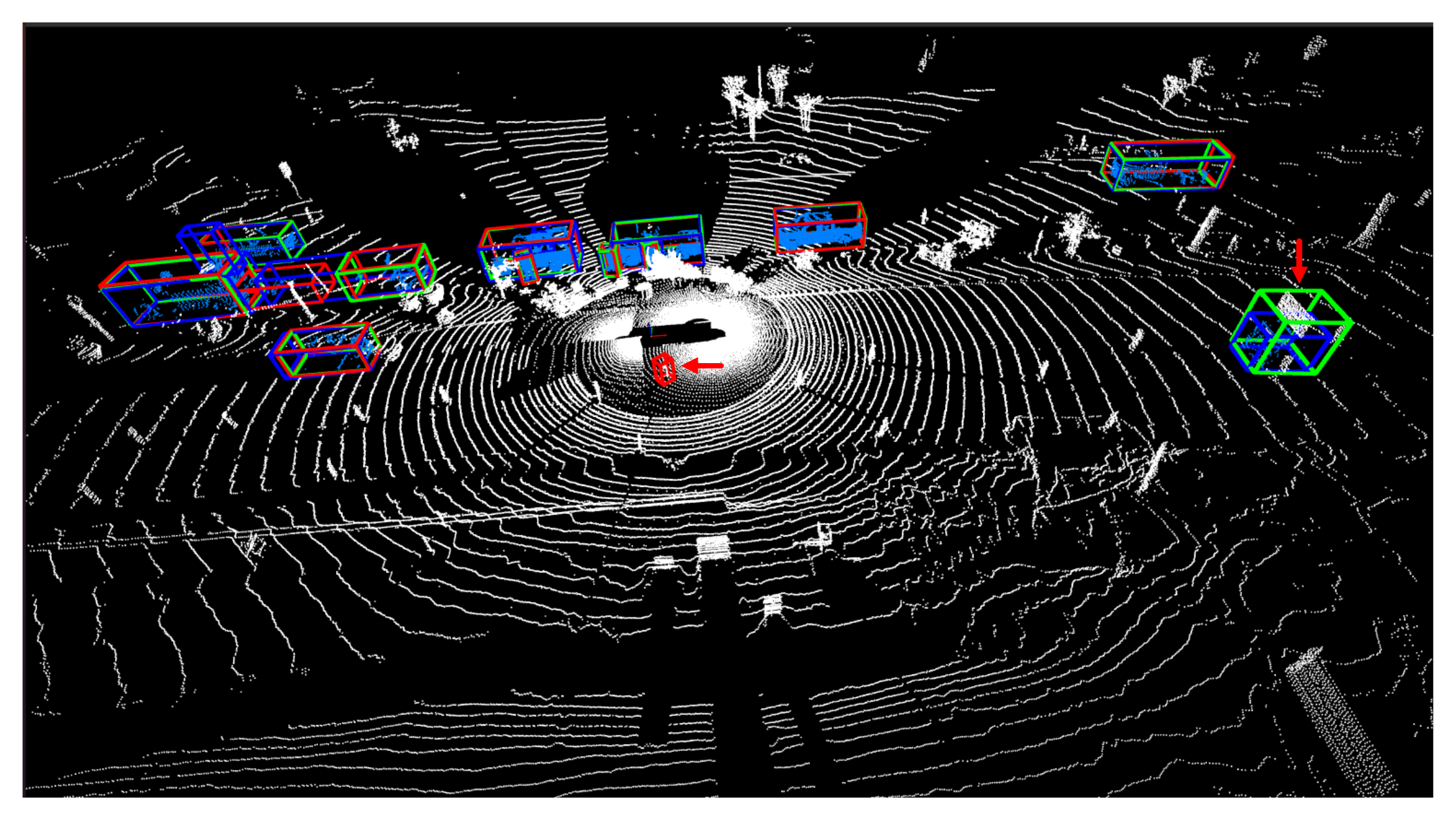} 
        \caption{Image(141)}
        \label{fig:image3}
    \end{subfigure}
    \hspace{0.02\textwidth}  
    \begin{subfigure}[t]{0.47\textwidth}
        \centering
        \includegraphics[height=5cm,width=\textwidth]{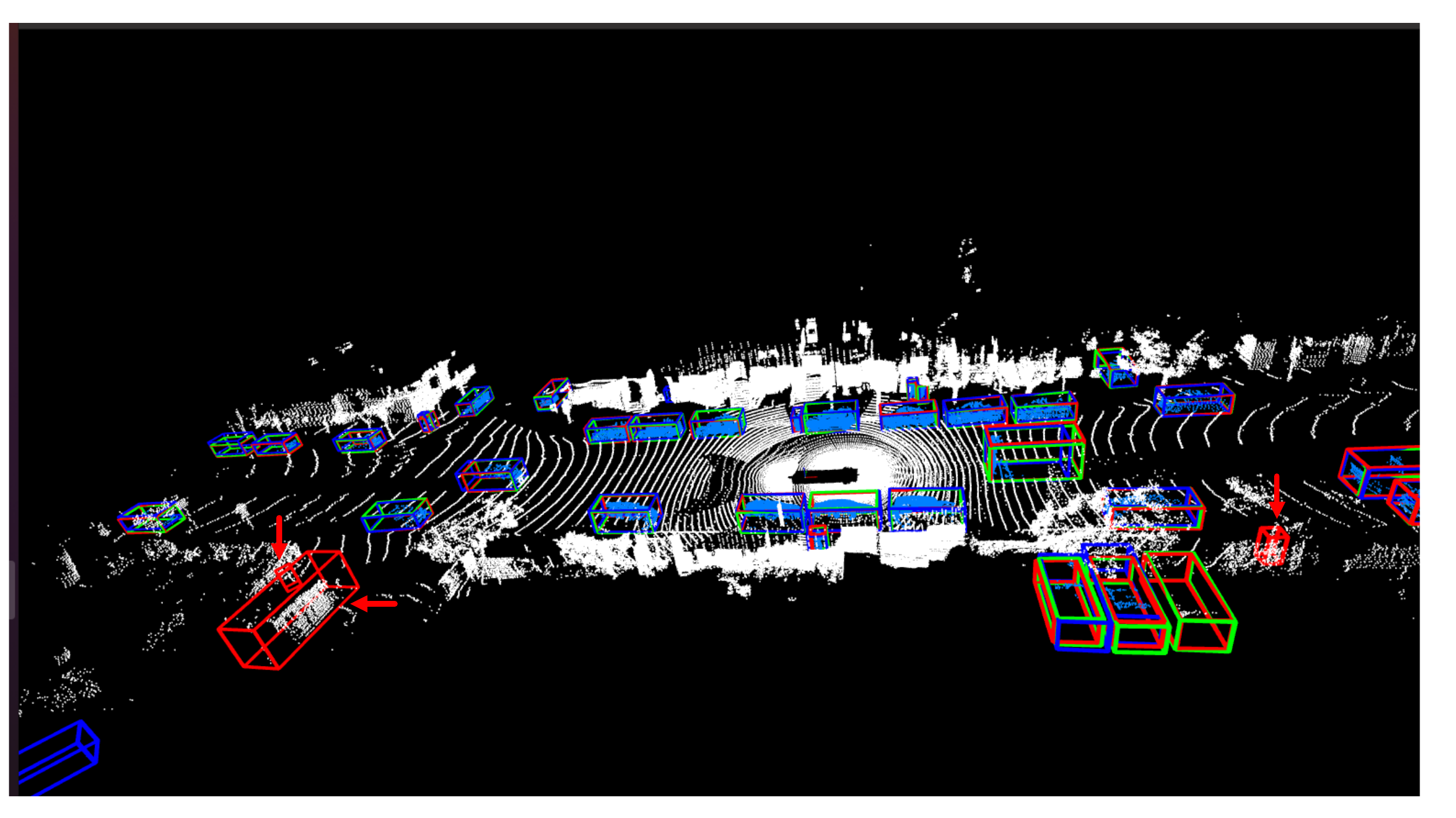} 
        \caption{Image(4066)}
        \label{fig:image4}
    \end{subfigure}
    \caption{Qualitative results on Waymo. The blue, red, and green boxes are human annotations, SAFDNet-Pillar predictions and PointSlice predictions, respectively. The blue points denote the points that fall within the human-annotated boxes. Compared to SAFDNet-pillar, PointSlice demonstrates superior accuracy in object identification and exhibits more effective detection in reducing false positives. In the point cloud visualization process, the score threshold is set to 0.3, and the non-maximum suppression (NMS) threshold is set to 0.1. Red arrows mark the prediction differences.}
    \label{fig:four_images}
\end{figure*}

As shown in the \cref{fig:four_images}, we compared the visualization results of PointSlice and SAFDNet-pillar on the Waymo validation dataset. The visualization results indicate that our model demonstrates superior performance in terms of detection accuracy, whereas the SAFDNet-pillar network exhibits issues with false positive object detections.

\section{Discussion and Conclusion}
\label{sec:conclusion}

In summary, to overcome the challenges of low inference efficiency in voxel-based models and reduced detection accuracy in pillar-based models, we developed a horizontal slicing approach for 3D point cloud data. This method effectively mitigates the computational burden associated with voxel-based models. Additionally, we introduced the Slice Interaction Network (SIN) to facilitate feature interaction across slices, resulting in the PointSlice model. Extensive evaluations on the Waymo, nuScenes, and Argoverse 2 datasets demonstrate that PointSlice achieves a superior trade-off between efficiency and accuracy, validating the effectiveness of our proposed approach.

\vspace{0.5em}
\noindent \textbf{Broader Impacts and Community Benefits.} 
Beyond the immediate performance improvements, our proposed slice-based paradigm offers a fresh perspective for the 3D perception community. By seamlessly bridging the gap between 3D voxel operations and 2D planar processing, PointSlice provides a highly efficient and easily deployable framework. Researchers and engineers in autonomous driving and mobile robotics can directly benefit from this architecture by adapting our 2D slice encoding to other resource-constrained 3D vision tasks, such as LiDAR-based multi-object tracking, multimodal sensor fusion, or real-time embodied AI applications.

\vspace{0.5em}
\noindent \textbf{Limitations and Future Directions.} 
Despite the promising efficiency of PointSlice, we acknowledge certain limitations. First, the fixed-height slicing strategy introduces quantization errors along the vertical axis, which may discard fine-grained geometric details essential for recognizing small objects like pedestrians. Second, while the 2D backbone ensures high inference speed, its feature representation capacity is inherently constrained compared to computationally intensive fully 3D voxel-based counterparts. Furthermore, although PointSlice achieves faster inference than voxel-based detectors, a performance gap in inference speed still exists compared to the most efficient pillar-based architectures.

Based on these observations, future research can explore several potential avenues for improvement. To address the quantization issue, developing adaptive or multi-scale slicing mechanisms could help preserve vertical details more effectively. Regarding scalability, investigating deeper backbones and advanced attention-based mechanisms within the Slice Interaction Network offers a promising path to further close the accuracy gap with state-of-the-art methods. Finally, designing more concise and efficient point cloud encoding strategies remains a key direction to further enhance inference efficiency.








\bibliographystyle{elsarticle-num}  
\bibliography{main}  
\end{document}